\pdfoutput=1
\documentclass[10pt,twocolumn,letterpaper]{article}

\usepackage[pagenumbers]{cvpr} 

\usepackage{booktabs}
\usepackage{times}
\usepackage{epsfig}
\usepackage{graphicx}
\usepackage{color}
\usepackage{lipsum}
\usepackage{xcolor}
\usepackage{commath}
\usepackage{colortbl}
\usepackage{wrapfig}
\usepackage[ruled]{algorithm2e}
\usepackage{enumerate}
\usepackage{xspace}
\usepackage{booktabs}
\usepackage{tabulary}
\usepackage{floatrow}
\usepackage{wrapfig}
\usepackage{appendix}
\usepackage[demo,abs]{overpic}
\usepackage{colortbl}
\usepackage{bm}
\usepackage{algpseudocode}

\usepackage{paralist}
\usepackage[export]{adjustbox}

\usepackage{soul}
\usepackage{subcaption}
\usepackage{graphicx}

\usepackage{amsmath, amssymb, caption, subcaption, multirow, textpos}

\newcommand{\nameofmethod}{AvatarGen}
\newcommand{\bb}[1]{\boldsymbol{\mathbf{#1}}}

\usepackage[pagebackref,breaklinks,colorlinks]{hyperref}

\usepackage[capitalize]{cleveref}
\crefname{section}{Sec.}{Secs.}
\Crefname{section}{Section}{Sections}
\Crefname{table}{Table}{Tables}
\crefname{table}{Tab.}{Tabs.}
\usepackage[inline]{enumitem}
\def\eg{\emph{e.g}\onedot} 
\def\ie{\emph{i.e}\onedot} 
 
 \def\vs{\emph{vs}\onedot}

\makeatletter
\DeclareRobustCommand\onedot{\futurelet\@let@token\@onedot}
\def\@onedot{\ifx\@let@token.\else.\null\fi\xspace}

\def\eg{\emph{e.g}\onedot} 
\def\ie{\emph{i.e}\onedot} 
 
 \def\vs{\emph{vs}\onedot}

\makeatother

\newcolumntype{x}[1]{>{\centering\arraybackslash}p{#1pt}}
\newcolumntype{y}[1]{>{\raggedright\arraybackslash}p{#1pt}}
\newcolumntype{z}[1]{>{\raggedleft\arraybackslash}p{#1pt}}
\newlength\savewidth

\DeclareMathOperator*{\argmin}{arg\,min}
\makeatother

\begin{document}

\title{\nameofmethod{}: A 3D Generative Model for  Animatable Human Avatars}

    \author{%
  Jianfeng Zhang$^{1*}$,
  Zihang Jiang$^{1*}$,
  Dingdong Yang$^2$,
  Hongyi Xu$^2$,
  \\
  Yichun Shi$^2$,
  Guoxian Song$^2$,
  Zhongcong Xu$^1$,
  Xinchao Wang$^1$,
  Jiashi Feng$^2$
  \\
  $^1$National University of Singapore \quad
  $^2$ByteDance
}

\twocolumn[{
\renewcommand\twocolumn[1][]{#1}

\begin{center}
\maketitle
    \captionsetup{type=figure}
    \includegraphics[width=0.95\textwidth]{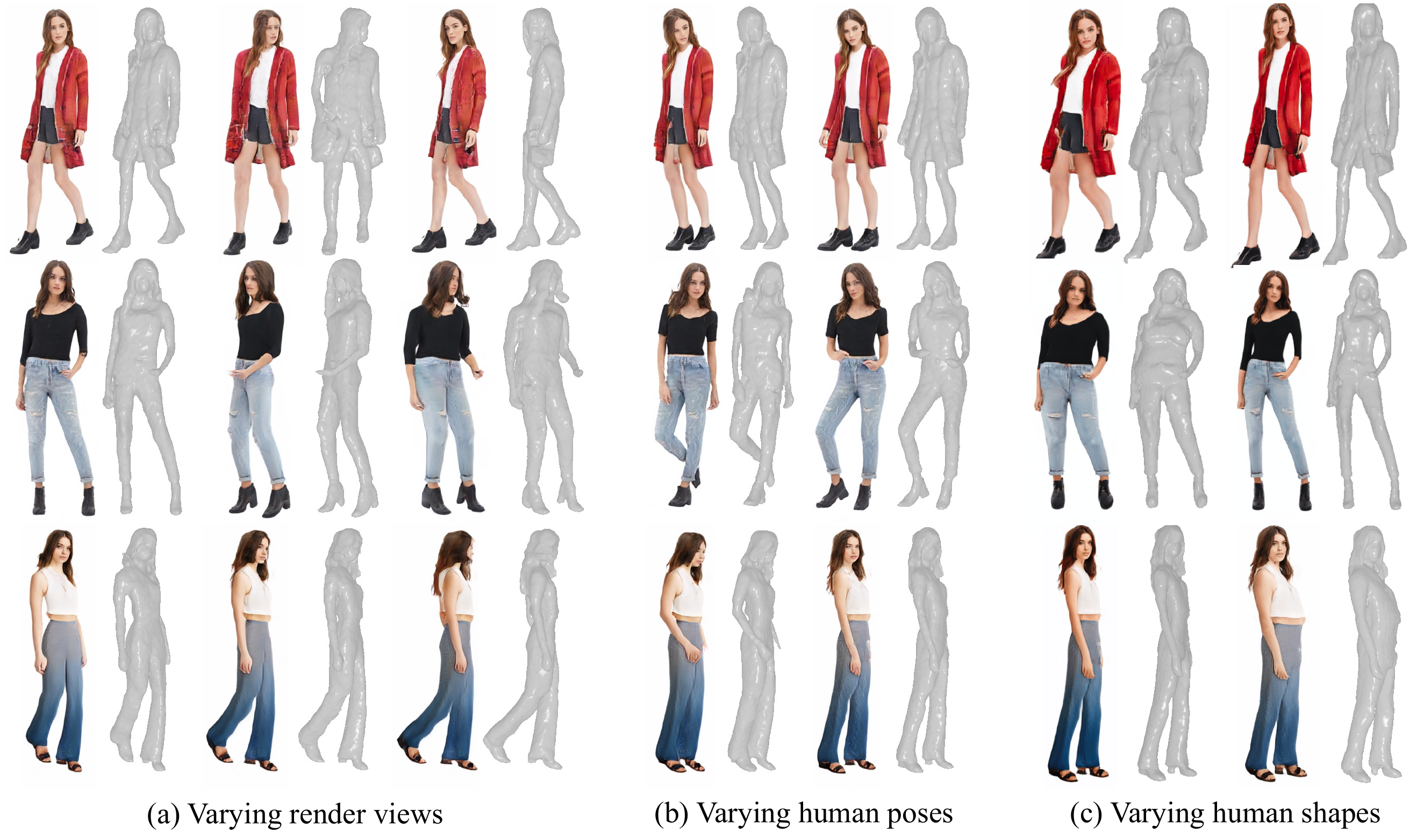}
    \captionof{figure}{Our \nameofmethod{}  can  synthesize  clothed 3D human avatars with detailed geometries and diverse appearances under disentangled control over camera viewpoints, human poses and shapes. Please see our \href{http://jeff95.me/projects/avatargen.html}{project page} for video results. }
    \label{fig:teaser}
\end{center}

}]

\let\thefootnote\relax\footnotetext{*Equal contribution.}

\begin{abstract}
\vspace{-3mm}
 Unsupervised generation of 3D-aware clothed  humans with various appearances and controllable geometries is important for creating virtual human avatars and other AR/VR applications. Existing methods are either limited to rigid object modeling, or not generative  and thus unable to generate high-quality virtual humans and animate them. In this work, we propose \nameofmethod{}, the first method that enables not only geometry-aware clothed human synthesis with high-fidelity appearances but also disentangled human animation controllability, while only requiring 2D images for training. Specifically, we decompose the generative 3D human synthesis into pose-guided mapping and canonical representation with predefined human pose and shape, such that the canonical representation can be explicitly driven to different poses and shapes with the guidance of a 3D parametric human model SMPL.
\nameofmethod{} further introduces a deformation network to learn non-rigid deformations for modeling fine-grained geometric details and pose-dependent dynamics. To improve the geometry quality of the generated human avatars, it leverages the signed distance field as geometric proxy, which allows more direct regularization from the 3D geometric
priors of SMPL. Benefiting from these designs, our method can generate animatable 3D human avatars with high-quality appearance and geometry modeling, significantly outperforming previous 3D GANs.
Furthermore, it is competent for many applications, \eg, single-view reconstruction, re-animation, and text-guided synthesis/editing. Our code and  models will be available at the \href{http://jeff95.me/projects/avatargen.html}{project page}.

\end{abstract}

\section{Introduction}

Generating  diverse and  high-quality 3D-aware virtual humans (avatars) with precise control over their geometries, \eg, poses and shapes, is a fundamental but extremely challenging task. Solving this task will benefit many applications like immersive photography visualization~\cite{zhang2022neuvv}, virtual  try-on~\cite{liu2021spatt}, VR/AR~\cite{xiang2021modeling,jiang2022selfrecon} and image editing~\cite{zhang2021editable,hong2022avatarclip}.

Conventional solutions rely on classical graphics modeling and rendering techniques~\cite{debevec2000acquiring,collet2015high,dou2016fusion4d,su2020robustfusion} to create virtual avatars. Though offering high-quality, they typically require pre-captured templates, multi-camera systems, controlled studios, and long-term works of artists. In this work, we aim to make virtual human avatars widely accessible at low cost. To this end, we propose the first 3D-aware avatar generative model  that can \emph{synthesize} 
\begin{inparaenum}[1)]
    \item high-quality virtual humans with 
    \item various appearances and disentangled geometry controllability
    \item and be {trainable from only 2D  images, thus largely alleviating the effort to create avatars.}
\end{inparaenum}

The 3D-aware generative models have recently seen rapid progress, fueled by introducing implicit neural representation (INR) methods~\cite{chen2019learning,Park_2019_CVPR,mescheder2019occupancy,mildenhall2020nerf} into generative adversarial networks \cite{chan2021pi,niemeyer2021giraffe,or2021stylesdf,gu2021stylenerf,chan2021efficient}.
However, these models are limited to  relatively simple and rigid objects, such as human faces and cars, and mostly fail to generate clothed human avatars whose appearance is highly sundry because of their articulated poses and great variability of clothing. Besides, they have limited control over the generation process and thus  cannot animate the generated objects, \ie, driving the objects to move by following certain instructions.
Another line of works leverage INRs~\cite{mildenhall2020nerf} to learn articulated human avatars for reconstructing a single subject from one's multi-view images or videos~\cite{peng2021animatable,2021narf,xu2021h,chen2021geometry,peng2022animatable}. While being animatable, these methods are \textit{not generative} and  thus cannot synthesize novel identities and appearances.

Targeting at generative modeling of animatable human avatars, we propose \nameofmethod{}, the first model that can synthesize \emph{novel} human avatars with disentangled control over their geometries and appearances. \nameofmethod{} is built upon  EG3D~\cite{chan2021efficient}, a recent   method that can synthesize    3D-aware {human faces}   via introducing  an effective tri-plane representation.  However,   it  is not directly applicable for human  avatar generation since it cannot handle the  challenges in modeling complex textures and   the articulated body structure with various  poses. Moreover, EG3D has limited control 
ability and thus it hardly animates the generated objects.

{To address these challenges, we propose to decompose the generative human avatar modeling into  \textit{pose-guided canonical mapping} and \textit{canonical human generation}. Guided by a 3D parametric human model  SMPL~\cite{loper2015smpl}, \nameofmethod{} un-warps each  point  sampled in the observation space with a specified human body to a standard avatar with predefined pose and shape, represented by tri-plane~\cite{chan2021efficient}, in a canonical space via  the inverse LBS~\cite{huang2020arch}.}
To accommodate  the non-rigid deformation between the observation and canonical spaces (like clothes wrinkle), our method further trains a deformation module to predict the proper residual deformation.
As such, it can generate {detailed geometry and texture for the observation space} by deforming the canonical one, which is much easier to generate and shareable  across different instances, thus largely alleviating the learning difficulties.
Meanwhile,  {such decomposed} formulation by design   \textit{disentangles  the geometries  and appearances}, offering   independent control over them.

Although our method can generate  3D human avatars with reasonable geometry, we find it tends to produce noisy body surfaces due to the lack of constraints on the learned geometry (density field). 
Inspired by recent works on neural implicit surface~\cite{wang2021neus,yariv2021volume,or2021stylesdf,peng2022animatable}, we propose to use the signed distanced field (SDF) {to impose stronger  \textit{geometry-aware guidance} for the model training}. 
With this, the model can leverage the prior body model to infer  reasonable signed distance values, which greatly improves the quality of the human avatar generation and animation.
We also introduce   a local face discriminator to alleviate undesirable face generation due to low-resolution faces in the training images. 

As shown in Fig.~\ref{fig:teaser}, trained from only  2D images,
\nameofmethod{}  can synthesize a large variety of clothed human avatars with high-fidelity appearances, geometries and disentangled controllability. 
We evaluate \nameofmethod{} quantitatively and qualitatively, demonstrating it strongly outperforms previous state-of-the-art methods.
Moreover, our \nameofmethod{} support several applications, like single-view 3D reconstruction, re-animation and text-guided synthesis.

\begin{figure*}[t]
    \centering
    \small
    \includegraphics[width=.9\linewidth]{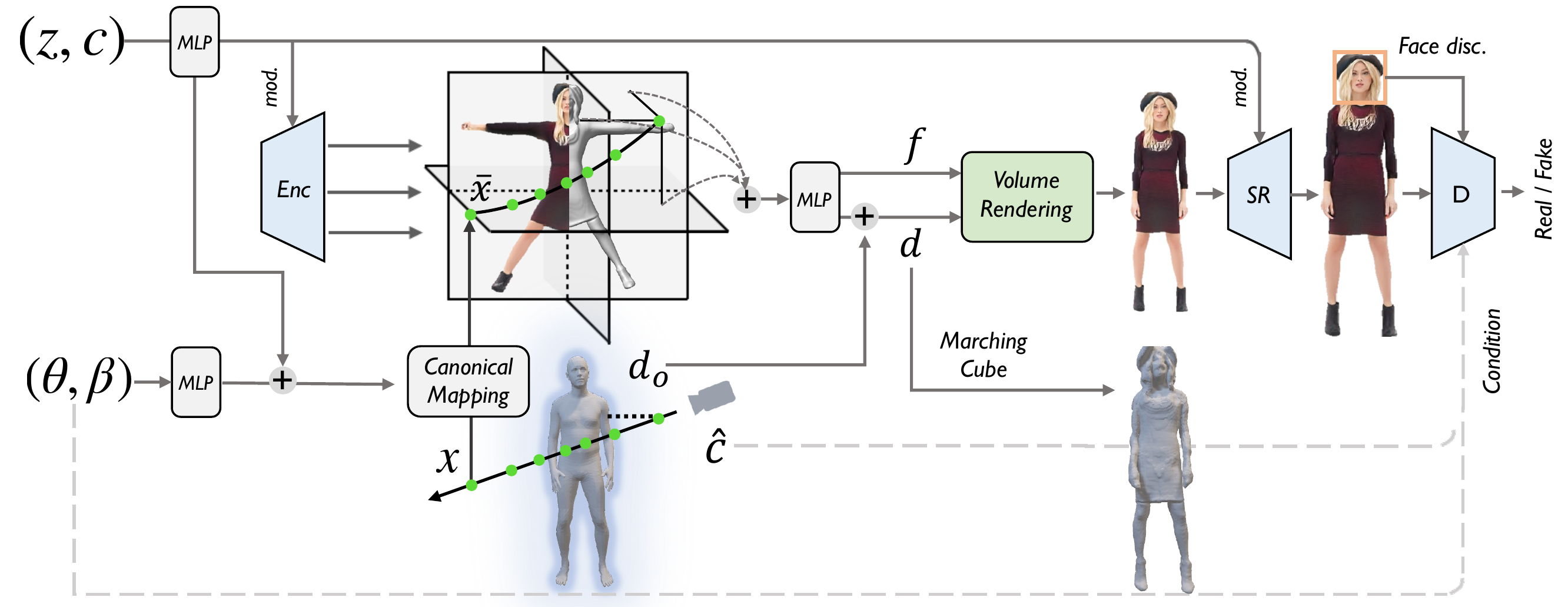}
    \caption{\textbf{Pipeline of \textit{\nameofmethod{}}}. Taking the latent code $\bb{z}$ and camera pose ${\mathbf{c}}$ as inputs, the encoder   generates   tri-plane features of a standard human avatar with canonical pose and shape. The geometry condition $(\bb{\theta}, \bb{\beta})$ are mapped together with latent code and applied to modulate the deformation module which un-warps the sampled points $\bb{x}$ from the observation space to the canonical space under the guidance of the SMPL model. The deformed positions $\Bar{\bb{x}}$ are   used to query features from the canonical tri-plane, which are then rendered as low-resolution features and images via the SDF-based volume renderer. Finally, a super-resolution module  decodes the feature images to high resolution images. The generator and the discriminator with camera and geometry conditions are optimized via adversarial training.
    }
    \label{fig:arch}
\end{figure*}
\section{Related Works}

\noindent {\bf Generative 3D-aware image synthesis.}
Generative adversarial networks (GANs)~\cite{goodfellow2014generative} have recently achieved photo-realistic image synthesis~\cite{karras2018progressive,karras2019style,Karras2020stylegan2,Karras2021}. Extending these capabilities to 3D settings has started to gain attention.
Early methods combine GANs with voxel~\cite{wu2016learning,hologan,nguyen2020blockgan}, mesh~\cite{Szabo:2019,Liao2020CVPR} or point cloud~\cite{achlioptas2018learning,li2019pu} representations for 3D-aware image synthesis.
Recently, several methods represent 3D objects by learning an implicit neural representation (INR)~\cite{schwarz2020graf,chan2021pi,niemeyer2021giraffe,chan2021efficient,or2021stylesdf,gu2021stylenerf,deng2021gram}. 
Among them, some methods use INR-based model as generator~\cite{schwarz2020graf,chan2021pi,deng2021gram}, while some others combine INR generator with 2D decoder for higher-resolution image generation~\cite{niemeyer2021giraffe,gu2021stylenerf,xue2022giraffe,zhang20223d}. 
Follow-up works like EG3D~\cite{chan2021efficient} proposes an efficient Tri-plane representation for  3D objects modeling and StyleSDF~\cite{or2021stylesdf} replaces density field with SDF for better geometry quality.
However, such methods are typically not easily extended to non-rigid humans due to  complex body articulation and appearance variations. 
Moreover, they have limited control over the generation process, making the generated objects hardly be animated. 
Differently, we study the problem of 3D-aware generative modeling of clothed humans, allowing disentangled control over the geometries and appearances.

\vspace{1mm}
\noindent  {\bf 3D human reconstruction and animation.}
Traditional human reconstruction methods require complicated hardware that is expensive for daily use, such as depth sensors~\cite{collet2015high,dou2016fusion4d,su2020robustfusion} or dense camera arrays~\cite{debevec2000acquiring,guo2019relightables}. 
To alleviate the requirement on the capture device, some methods train neural networks to reconstruct human models from  images with differentiable renderer~\cite{xu2021texformer,Gomes2022}. 
Recently, neural radiance fields (NeRF)~\cite{mildenhall2020nerf} employ the volume rendering to learn density and color fields from dense camera views. Some methods augment NeRF with human body priors to enable 3D human reconstruction from sparse multi-view data~\cite{peng2021neural,chen2021geometry,xu2021h,su2021anerf}.
Follow-up improvements~\cite{peng2021animatable,chen2021animatable,liu2021neural,peng2022animatable,weng2022humannerf} are made by combining implicit representation with the SMPL model and exploiting the linear blend skinning techniques to learn animatable 3D human modeling from temporal data. 
However, these methods are not generative, \ie, they cannot synthesize novel identities and appearances. 
{The concurrent works ENARF-GAN~\cite{noguchi2022unsupervised} and EVA3D~\cite{EVA3D} also leverage 3D human priors to learn animatable 3D human GAN. However, they still suffer from undesirable  artifacts and lack of precise geometry controls. Instead, our \nameofmethod{} achieves high-fidelity  human image synthesis with disentangled geometry controllability.}

\section{Methodology}

\subsection{Overview}

{We introduce a 3D generative model $G$, named AvatarGen, that can synthesize  high-quality multi-view consistent human images, {with disentangled controllability on   the human   geometries (\eg, pose and shape)}. It only requires    2D images for training,  without using multi-view, temporal information or 3D human scan annotations.}

Fig.~\ref{fig:arch} illustrates  the pipeline of our proposed AvatarGen.  Given a randomly sampled latent code $\bb{z}$ from Gaussian distribution,  conditioning camera  $\bb{c}$, 
and a geometry parameter $\bb{p}=(\bb{\theta},\bb{\beta})$, where $\bb{\theta}$  and $\bb{\beta}$ specify human pose and shape respectively, 
its generator $G$ produces a 3D neural representation (\ie, tri-plane~\cite{chan2021efficient} as detailed later) and synthesizes  a photo-realistic  human image  $I=G(\mathbf{z}|\mathbf{c},\mathbf{p})$ with the corresponding viewpoint, pose and shape. In this work, we use 3D parametric model   SMPL~\cite{loper2015smpl} to represent the   human geometries $\mathbf{p}$. Following EG3D~\cite{chan2021efficient}, we associate each training image with a set of camera  $\mathbf{c}$ and SMPL parameters $\mathbf{p}$ estimated from  an off-the-shelf pose estimator~\cite{kocabas2021pare}.

{We choose tri-plane~\cite{chan2021efficient} as the representation for human avatars because it is efficient and effective for high-fidelity image synthesis. 
The generator   encodes $\mathbf{z}$ and camera condition ${\bb{c}}$  to a  tri-plane feature field.
Then given a desired camera pose $\hat{\bb{c}}$, a high-dimensional feature map is synthesized via querying the tri-plane features  and integrating MLP-based neural radiance fields (color feature $\bb{f}$ and density ${\sigma}$) along the camera rays. By feeding this feature map to a super-resolution module~\cite{Karras2020stylegan2}, the generator produces   the final image $I$ at high resolution.} We optimize $G$ with a discriminator module $D$ via adversarial training, resulting in view-consistent image synthesis.

{To disentangle human pose and shape from appearance synthesis and achieve controllable generation, we propose a decomposition of a 3D human  representation. Specifically, we model a neural human avatar in a canonical space with predefined pose and shape, while we control the final image synthesis with desired geometries by deforming the sampled rays according to the   geometry parameters $\bb{p}$ (Sec.~\ref{sec:decompose}). }
In contrast to previous works that use density fields for 3D-aware GAN~\cite{chan2021efficient,gu2021stylenerf,niemeyer2021giraffe}, we leverage signed distance fields (SDFs) as geometry proxy, which offers  more direct geometric regularization (Sec.~\ref{sec:sdf}).
We optimize our generator $G$  with  a whole body and a local head discriminators via adversarial training and several carefully designed geometric regularization (Sec.~\ref{sec: training}).

\subsection{Animatable 3D-aware Human  GAN}
\label{sec:decompose}

There are two main challenges for generating 3D-aware human avatars with controllable geometries. The first is how to effectively integrate specified geometry condition  $\bb{p}$ into the generative  model, making the generated avatar   controllable and animatable.  
The second challenge is how to guarantee the generation quality, considering  learning pose-dependent clothed human appearance and geometry 
is highly under-constrained and the model training is difficult. To tackle these challenges, inspired by recent dynamic neural rendering works~\cite{peng2021animatable,park2021nerfies,xu2021h}, we propose to decompose the  human avatar generation into two steps: \textit{pose-guided canonical mapping} and \textit{canonical human generation}.

\vspace{1mm}
\noindent  {\bf Pose-guided canonical mapping.}
 We define the human 2D image with SMPL  $\bb{p}$ as the \textit{observation} space $\mathcal{O(\bb{p})}$. To relieve learning difficulties, our model attempts to deform the observation space  to a \textit{canonical} space $\mathcal{C(\bar{\bb{p}})}$ with a predefined template SMPL body $\bar{\bb{p}}$ that is shared across different identities. The deformation function $T: \mathbb{R}^3 \mapsto \mathbb{R}^3$ thus maps any spatial points $\mathbf{x}_i$ sampled in the observation space to $\bar{\bb{x}}_i$ in the canonical space for volumetric rendering. 

Learning such a deformation function has been proved effective for dynamic scene modeling~\cite{park2021nerfies,pumarola2021d}. However, learning to deform in such an implicit manner cannot  handle large articulation of humans and thus hardly generalizes to novel poses/shapes. To overcome this limitation, we use SMPL to explicitly guide the deformation~\cite{liu2021neural,peng2021animatable,chen2021animatable}.
SMPL defines a skinned vertex-based human model   $(\mathcal{V},\mathcal{S})$, where $ \mathcal{V} = \{\mathbf{v} \} \in \mathbb{R}^{N \times 3}$ is the set of $N$  vertices  and $  \mathcal{S} = \{\bb{s}\} \in \mathbb{R}^{N \times J}$ is a set of the skinning weights  assigned for the vertex w.r.t.\ $J$ joints, with  $\sum_j s_j=1, s_j \geq 0$ for every joint.
With inverse LBS (IS) transformation~\cite{jacobson2012fast}, we can map the SMPL body in the observation space with parameters $\mathbf{p}$ into the canonical space as:
\begin{equation}
    T_{\text{IS}} (\bb{v}, \bb{s}, \bb{p}) = \sum_j s_j \cdot (R_j\mathbf{v}+\mathbf{t}_j),
\end{equation}
where $R_j$ and $\mathbf{t}_j$ are the rotation and translation at each joint $j$ derived from  SMPL  with $\mathbf{p}$.

This formulation can be  extended to any spatial points in the observation space by  adopting the same transformation from the nearest point on the surface of SMPL body~\cite{huang2020arch}. Formally,
for any  points $\bb{x}_i$, we first find its nearest point $\bb{v^{*}}$ on the SMPL body surface as $\bb{v^{*}}=\argmin_{\bb{v}\in \mathcal{V}}||\bb{x}_i-\bb{v}||_2$. 
Then, we use the corresponding skinning weights $\bb{s^*}$ to deform $\bb{x}_i$ to $\bar{\bb{x}}_i'$ in the canonical space as:
\begin{equation}
    \bar{\bb{x}}_i' = T_o (\bb{x}_i| \bb{p}) = T_{\text{IS}} ({\bb{x}_i}, \mathbf{s^*}, \bb{p}).
    \label{eqn:invskin}
\end{equation}

Although the SMPL-guided inverse LBS can help align the rigid skeletons with the template body, it lacks the ability to model the pose/shape-dependent deformation, like cloth wrinkles.
To alleviate this issue, we further train a deformation network to learn a residual deformation that completes the fine-grained geometric deformation by
\begin{equation}
\begin{aligned}
   \Delta \bar{\bb{x}}_i = \text{MLPs}(\text{Concat}[\text{Embed}(\mathbf{x}_i),\mathbf{w},\mathbf{p}]),
\end{aligned}
\end{equation}
where $\mathbf{w}$ is the latent style code mapped from the input latent code $\mathbf{z}$ via MLP, which contains appearance and geometric details of the current generation. We concatenate it with the position-embedded $\mathbf{x}_i$ and SMPL $\mathbf{p}$ and feed them to MLPs to yield the residual deformation. 
Thus, the final canonical mapping $T$ is formulated as
\begin{equation}
\begin{aligned}
    \bar{\bb{x}}_i = T(\mathbf{x}_i)  =\bar{\bb{x}}_i'+\Delta \bar{\mathbf{x}}_i
\end{aligned}
\end{equation}

\vspace{1mm}
\noindent  {\bf Canonical human generation.}
{After  warping 3D points in $\mathcal{O(\bb{p})}$ back to $\mathcal{C(\bar{\bb{p}})}$, \nameofmethod{} leverages the tri-plane representation for 3D-aware generation of clothed humans in the canonical space.}
More concretely, it first generates a canonical tri-plane via a StyleGAN backbone by taking the latent code $\mathbf{z}$ and camera parameters $\mathbf{c}$ as inputs. Then, for each point $\bb{x}$ back-warped to $\bar{\bb{x}}$, the model queries the canonical tri-plane features followed by  MLPs to decode color-based features $\bb{f}$ and density $\sigma$ for volume rendering and the following super-resolution. Note that our canonical tri-plane only needs to generate appearance and geometry with a  canonical pose and shape, which alleviates the optimization difficulties and substantially helps the learning of high-quality human generation with disentangled controllability.

\subsection{Geometry-aware Human Modeling}
\label{sec:sdf}
To improve  geometry modeling quality of \nameofmethod{}, inspired by recent neural implicit surface works~\cite{wang2021neus,yariv2021volume,or2021stylesdf,peng2022animatable}, we adopt signed distance field (SDF) instead of density field as geometry proxy, because it introduces  more direct regularizations~\cite{or2021stylesdf,peng2022animatable} and allows the model fully leverage geometric knowledge of the 3D SMPL model as guidance.

\vspace{1mm}
\noindent  {\bf SDF prediction with geometric prior.}
Directly adopting SDF 
for clothed human modeling is non-trivial due to the complicated body articulation, pose-dependent deformation and insufficient supervisions from  2D images. 

To solve these issues, we propose a simple yet effective SDF prediction scheme that fully exploits geometric priors of the SMPL body model.
Specifically, given a desired SMPL parameter $\bb{p}=(\bb{\theta}, \bb{\beta})$, we obtain its corresponding body mesh in the observation space as $M=T_\text{SMPL}(\bb{p})$, where $T_\text{SMPL}$ is the SMPL transformation function. Then, for each 3D point $\mathbf{x}$, instead of predicting its signed distance value directly, we first compute its coarse signed distance $d_o(\bb{x}|\bb{p})$ to the closest surface point of $M$. Then, we feed $d_{o}$ alone with the  features sampled from tri-plane to a light-weight MLP to predict the \textit{residual} signed distance $\Delta d$. The residual SDF models the fine-grained surface details, such as hairs, clothes and wrinkles that are not represented by the SMPL model (Fig.~\ref{fig:arch}). The final signed distance of each point is computed as $d(\bb{x}|\bb{z},\bb{c},\bb{p})=d_{o}+\Delta d$. Predicting SDFs on top of the coarse body mesh largely alleviating the geometry learning difficulties, thus achieving better generation and animation results. Please refer to Appendix for more details.

Moreover, despite the disentangled pipeline as illustrated in Sec.~\ref{sec:decompose}, there is no explicit loss that constrains the generated geometry to be consistent with the human pose and shape defined by $\bb{p}$. Therefore, we guide the predicted SDF ${d}$ by minimizing its difference to the SDF ${d}_o$ defined by the underlying SMPL body as 
\begin{equation}
\label{eqn:prior}
    \begin{split}
    L_{prior} &= \frac{1}{|R|} \sum_{\bb{x}\in R} w(\bb{x} |  \bb{p}) || {d}(\bb{x} | \bb{z}, \bb{c}, \bb{p}) - {d}_o(\bb{x} | \bb{p}) ||, \\
    & w(\bb{x} | \bb{p}) = \exp\left({\frac{-{d}_o(\bb{x}|\bb{p})^2}{\kappa}}\right).
\end{split}
\end{equation}
Here $R$ is the ray samples for the volume rendering, $w$ is the weight of prior loss for point $\bb{x}$ and $\kappa$ is a constant scalar that defines the tightness around the surface boundary.
Such a prior loss guides the generated geometry to follow the geometric attributes controlled by $\bb{p}$. 
We decay the weights of the geometric prior loss $L_{prior}$ as the point moving away from the SMPL body surface, allowing higher degrees of freedom in generation of residual geometries, such as hairs and clothes.

\vspace{1mm}
\noindent  {\bf SDF-based volume rendering.}
Following~\cite{or2021stylesdf}, we adopt SDF-based volume rendering to obtain the feature images. 
For any sampled points ${\bb{x}}$ on the rays, we  first un-warp it  to $\bar{\bb{x}}$ by canonical mapping. We query feature $F(\bar{\bb{x}})$ for $\bar{\bb{x}}$ from the canonical tri-plane, and  feed it into two MLPs to predict the color feature $\bb{f}=\text{MLP}_f(F(\bar{\bb{x}}))$ and the signed distance $d=d_o+\Delta d=d_o + \text{MLP}_{d}(F(\bar{\bb{x}}), d_o)$, where $d_o$ is body SDF queried from SMPL for ${\bb{x}}$. 
We then convert the signed distance value $d_i$ of each point $\mathbf{x}_i$ alone a ray $R$ to density value $\sigma_i$ as
$
\sigma_i = \frac{1}{\alpha}\cdot \text{Sigmoid}(\frac{-d_i}{\alpha}),
$
where $\alpha >0$ is a learnable parameter that controls the tightness of the density around the surface boundary.
By integration along the ray $R$ we can get the corresponding pixel feature as 
\begin{equation}
    I(R) = \sum_{i=1}^N \left(\prod_{j=1}^{i-1} e^{-\sigma_j \cdot \delta_j}\right) \cdot \left(1-e^{-\sigma_i \cdot \delta_i}\right) \cdot f_i,
\end{equation}
where $\delta_i = || \bb{x}_i-\bb{x}_{i-1} ||$ and $N$ is number of points sampled per ray. By aggregating all rays, we can get the final image feature which is  feed into a super-resolution module~\cite{Karras2020stylegan2} to generate the final high-resolution synthesized image.

\subsection{Training}
\label{sec: training}

We use the non-saturating GAN loss $L_{\text{GAN}}$~\cite{Karras2020stylegan2} with R1 regularization $L_{\text{Reg}}$~\cite{mescheder2018training} to train our model  end-to-end. We also adopt the dual-discriminator proposed by EG3D~\cite{chan2021efficient}. It feeds both the rendered raw image and the decoded high-resolution image into the discriminator 
for improving   consistency of the generated  multi-view   images. 
Moreover, as observed in ~\cite{Fruehstueck2022InsetGAN}, most of the human generation artifacts appear in face regions due to low resolution compared with full bodies. Thus, we further enhance the quality of the generated human faces by cropping the head regions and feeding it to an additional discriminator which is jointly trained with the whole framework. For $512^2$ resolution image, an $80^2$ square covering the whole head region is used. 
To obtain better controllability, we feed both SMPL parameters $\mathbf{p}$ and camera parameters $\hat{\mathbf{c}}$ as conditions to the discriminator for adversary training.
To regularize the learned SDFs, we apply Eikonal loss to the sampled points as:
\begin{equation}
    L_{\text{Eik}} = \sum_{\mathbf{x}_i} (||\nabla d_i|| -1)^2,
\end{equation}
where $\mathbf{x}_i$ and $d_i$ denote the sampled point  and predicted signed distance value, respectively. Following~\cite{or2021stylesdf}, we  adopt a minimal surface loss to encourage the model to represent human geometry with minimal volume of zero-crossings that penalizes  the SDF close to zero:  $L_{\text{Minsurf}} = \sum_{\mathbf{x}_i}\exp(-100 |d_i|).$
To prevent the learned deformation network from collapsing, we use a deformation regularization loss to regularize the residual deformation $\Delta \bar{\mathbf{x}}_i$ to be small: $L_{\text{Deform}} = \sum_{\mathbf{x}_i} ||\Delta \bar{\mathbf{x}}_i||$.
Along with the geometric prior loss in Eqn.~(\ref{eqn:prior}), the overall loss is  formulated as
\begin{equation}
\begin{aligned}
    L_{\text{total}} =& L_{\text{GAN}}+\lambda_{\text{Reg}}L_{\text{Reg}}+\lambda_{\text{Eik}}L_{\text{Eik}}+\lambda_{\text{Minsurf}}L_{\text{Minsurf}}
    \\& +\lambda_{\text{Deform}}L_{\text{Deform}}+\lambda_{\text{Prior}}L_{\text{Prior}},
\end{aligned}
\end{equation}
where $\lambda_{*}$ are the corresponding loss weights.

\begin{table*}[t]
    \centering
    \renewcommand{\tabcolsep}{3pt}\vspace{-2mm}
    \resizebox{1\textwidth}{!}{
    \begin{tabular}{@{\hskip 1mm}l c c c c| c c c c| c c c c | c c c c@{\hskip 1mm}}
	\toprule
	& \multicolumn{4}{c|}{MPV} & \multicolumn{4}{c|}{UBC} & \multicolumn{4}{c|}{DeepFashion} & \multicolumn{4}{c}{SHHQ}\\
    & FID $\!\downarrow$          & FaceFID $\!\downarrow$ & Depth $\!\downarrow$ & PCK $\!\uparrow$   & FID $\!\downarrow$  & FaceFID $\!\downarrow$ & Depth $\!\downarrow$ &  PCK $\!\uparrow$ & FID $\!\downarrow$  & FaceFID $\!\downarrow$ & Depth $\!\downarrow$  & PCK $\!\uparrow$ & FID $\!\downarrow$  & FaceFID $\!\downarrow$ & Depth $\!\downarrow$  & PCK $\!\uparrow$ \\
    \midrule
    StyleNeRF-$256^2$     & 10.71   & 24.32  & 1.46      & -        & 20.65  & 34.30 & 1.44 &   -    & 15.93  & 29.22    & 1.43 &   -   & 13.52 & 26.21 & 1.37 & -   \\
    StyleSDF-$512^2$	  & 43.79   & 69.71	& 1.79 	  	& -     & 34.12  & 39.05 & 1.80 &  -  & 45.06  & 41.78    & 1.77 & -  & 43.24 & 47.46 & 1.78 & - \\
    EG3D-$512^2$          & 15.44   & 33.98	& 1.31  	& - 	   & 14.55  & 20.24 & 1.28 &  -  & 14.36 & 34.99    & 1.44 & -  & 9.33 & 35.10 & 1.43 & -  \\
    \midrule
    ENARF-$128^2$	      & 75.10   & 57.64  & 2.32      	& 58.92 	   & 41.75     & 36.34 & 2.10    &    54.40   & 67.96 & 51.17    &   2.57  & 55.06 & 73.53 & 49.35 & 2.43 & 53.44   \\
    EVA3D-$512^2$         & -      & -  & -         & -        & 12.61 & - & -    & 99.17  & 15.91 &  -   & -    & 87.50  & 11.99 & - & -    & 88.95  \\
    \midrule
    \nameofmethod{}-$512^2$ 			& \textbf{5.25} 				    	& \textbf{6.89} 	            & \textbf{.429} 					& \textbf{98.79} 				    & \textbf{6.71} & \textbf{8.61} & \textbf{.453} & \textbf{99.38}   & \textbf{7.68} & \textbf{8.76} & \textbf{.433} & \textbf{99.24} & \textbf{4.29} & \textbf{4.51} & \textbf{.365} & \textbf{99.49} \\
		\bottomrule
    \end{tabular}
    }
    \caption{\textbf{Quantitative comparisons} with baselines on four datasets, with best results in bold.
    }
    \label{tab:main_results}
\end{table*}
\vspace{1mm}

\section{Experiments}
\label{experiments}

We evaluate  methods of 3D-aware clothed human generation on four real-world fashion datasets: MPV~\cite{dong2019towards}, UBC~\cite{zablotskaia2019dwnet}, DeepFashion~\cite{liuLQWTcvpr16DeepFashion} and SHHQ~\cite{fu2022styleganhuman}. They contain single clothed people in each image.
Since we focus on foreground human  generation, we use a segmentation model~\cite{paddleseg2019} to remove irrelevant backgrounds.
We adopt an off-the-shelf pose estimator~\cite{kocabas2021pare}  to obtain approximate camera and SMPL parameters.
We filter out images with partial observations and those with poor SMPL estimations, and get nearly 13K, 31K, 12K and 39K full-body images for each dataset, respectively.
We align and crop images according to the estimated keypoints  and scales, and resize them to $512^2$ resolution. 
We sample 48 points alone each ray for volumetric rendering.
Horizontal-flip augmentation is used during training.
We note that these datasets are primarily composed of front-view images—few images captured from side or back views.
To compensate this, we sample more side- and back-view images to re-balance viewpoint distributions following~\cite{chan2021efficient}. 
Our data processing scripts and pre-processed datasets will be released. 

\subsection{Comparisons}
\label{sec:mainresults}

\noindent  {\bf Baselines.}
We compare  \nameofmethod{} against five state-of-the-art methods for 3D-aware image synthesis: EG3D~\cite{chan2021efficient}, StyleSDF~\cite{or2021stylesdf}, StyleNeRF~\cite{gu2021stylenerf}, ENARF-GAN~\cite{noguchi2022unsupervised} and EVA3D~\cite{EVA3D}. 
EG3D and StyleNeRF combine volume renderer with 2D decoder for high-resolution image synthesis. StyleSDF uses SDF  for regularized geometry modeling. ENARF-GAN and EVA3D  leverage 3D human priors for animatable human generation as well. 

\noindent  {\bf Quantitative evaluations.}
Tab.~\ref{tab:main_results} provides quantitative comparisons between  \nameofmethod{} and  baselines. We measure image quality with Fr\'echet Inception Distance ({FID})~\cite{heusel2017gans} between 50k generated images and all of the available real images.  We also evaluate the quality of the generated faces by cropping the face regions ($80^2$ for $512^2$-resolution image) from the generated and real images to compute FaceFID.
We evaluate geometry quality by calculating Mean Squared Error  (MSE) against pseudo groundtruth (GT) depth-maps (\textit{Depth}) that are estimated from the generated  images by~\cite{saito2020pifuhd}. Following~\cite{noguchi2022unsupervised}, we use Percentage of Correct Keypoints (PCK) to evaluate the effectiveness of the pose controllability.
For additional evaluation details, please refer to the appendix. 
From Tab.~\ref{tab:main_results}, we observe our model outperforms all the baselines w.r.t.\ all the metrics and datasets. Notably, it outperforms baseline models by  significant margins (50.9\%, 51.1\%, 51.7\%, 54.0\% in FID) on four datasets. These results clearly demonstrate its superiority in clothed human synthesis. Moreover, it  maintains state-of-the-art geometry quality and pose accuracy.

\begin{figure*}[h]
    \centering
    \small
    \includegraphics[width=0.95\linewidth]{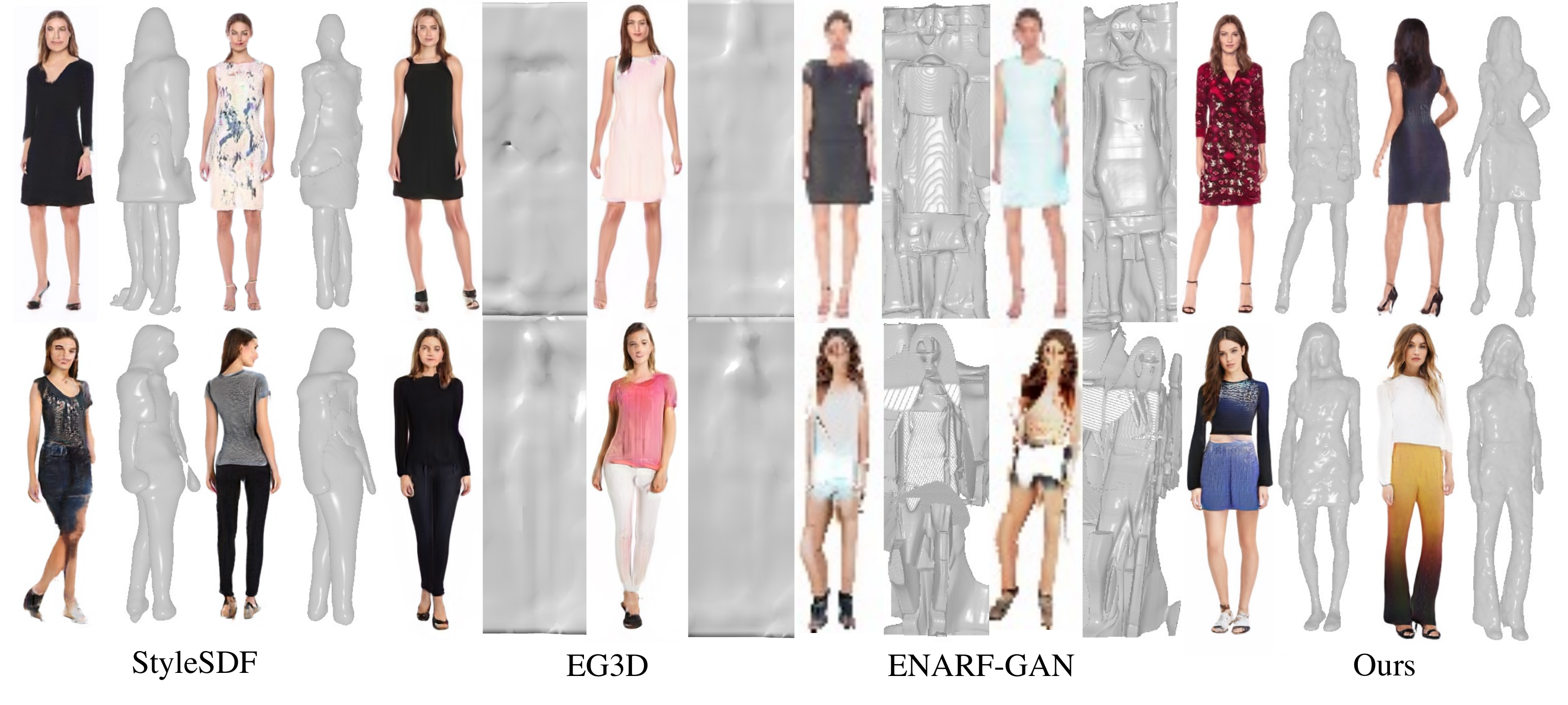}
 \caption{  \textbf{Qualitative comparisons} of generation results against baselines including StyleSDF~\cite{or2021stylesdf}, EG3D~\cite{chan2021efficient} and ENARF-GAN~\cite{noguchi2022unsupervised}.
 } 
    \label{fig:compare}
\end{figure*}

\vspace{1mm}
\noindent  {\bf Qualitative results.}
We show a qualitative comparison against baselines in Fig.~\ref{fig:compare}.  It can be observed that compared with our method, StyleSDF~\cite{or2021stylesdf} generate 3D avatar with over-smoothed geometry. In addition, the noises and holes can be observed around the generated avatars and the geometry details like face and clothes are missing.  EG3D~\cite{chan2021efficient} struggles to learn  3D human geometry from 2D images and suffers   degenerated qualities. Although using pose priors for 3D human generations, ENARF-GAN still suffers low rendering quality and undesired geometry.
Compared with them,  \nameofmethod{}  generates 3D-aware avatars with high-fidelity appearances and learns the complex human geometries from 2D images only. Also, \nameofmethod{} can generate avatar in diverse poses with much better geometric details. Please refer to our \href{http://jeff95.me/projects/avatargen.html}{project page} and Appendix for more results.

\begin{table*}[h]
	\renewcommand{\tabcolsep}{2pt}\vspace{-2mm}
	\begin{subtable}[!t]{0.33\textwidth}
		\centering
		\begin{tabular}{cccccccc}
			\toprule
			\textit{Geo.} & FID & FaceFID & Depth \\
			\midrule
			Density  & 7.87 & 11.21 & .652\\
			SDF  & 6.54 & 8.74 & .576\\
			\bottomrule
		\end{tabular}
		\caption{The effect of different geometry proxies.
		}
		\label{ablation:geoproxy}
	\end{subtable}
	\hspace{\fill}
	\begin{subtable}[!t]{0.34\textwidth}
		\centering
		\begin{tabular}{cccccccc}
			\toprule
			\textit{Deform.} & FID & FaceFID & Depth \\
			\midrule
			IS & 7.89 & 12.90 & .752\\
			IS+RD & 6.54 & 8.74 & .576\\
			\bottomrule
		\end{tabular}
		\caption{Deformation schemes.
		}
		\label{ablation:deformation}
	\end{subtable}
	\hspace{\fill} 
	\begin{subtable}[!t]{0.3\textwidth}
		\centering
		\begin{tabular}{cccccccc}
			\toprule
			\textit{Face Disc.} & FID & FaceFID & Depth \\
			\midrule
			w/o & 6.02 & 12.31 & .604\\
			w/ & 6.54 & 8.74 & .576\\
			\bottomrule
		\end{tabular}
		\caption{The effect of face discriminator. 
		}
		\label{ablation:facedisc}
	\end{subtable}
	\hspace{\fill}
	\begin{subtable}[b]{0.33\textwidth}
		\centering
		\begin{tabular}{cccccccc}
			\toprule
			\textit{SDF Scheme} & FID & FaceFID & Depth \\
			\midrule
			
			W/o & 7.16 & 10.63 & .689\\
			Can. & 7.09 & 9.27 & .592 \\
			Obs.+raw & 7.03 & 10.23 & .602\\
			Obs. & 6.54 & 8.74 & .576\\
			\bottomrule
		\end{tabular}
		\caption{SDF prediction schemes.
		}
		\label{ablation:bodysdf}
	\end{subtable}
	\hspace{\fill}
	\begin{subtable}[b]{0.34\textwidth}
		\centering
		\begin{tabular}{cccccccc}
			\toprule
			\textit{Ray Steps} & FID & FaceFID & Depth \\
			\midrule
			12 & 7.76 & 12.44 & .656\\
			24 & 7.59 & 12.02 & .636\\
			36 & 7.07 & 11.30 & .626\\
			48 & 6.54 & 8.74 & .576\\
			\bottomrule
		\end{tabular}
		\caption{Number of ray steps.
		}
		\label{ablation:ray_step}
	\end{subtable}
	\hspace{\fill}
	\begin{subtable}[b]{0.3\textwidth}
		\centering
		\begin{tabular}{ccccccccc}
			\toprule
			\textit{KNN} & FID & FaceFID & Depth \\
			\midrule
			1 & 6.54 & 8.74 & .576\\
			2 & 6.49 & 9.55 & .583 \\
			3 & 6.20 & 9.72 & .578 \\
			4 & 6.17 & 8.88 & .575 \\
			\bottomrule
		\end{tabular}
		\caption{KNN in inverse skinning deformation. 
		}
		\label{ablation:knn}
	\end{subtable}
	\hspace{\fill}
	\caption{\textbf{Ablation study} on Deepfashion. In (d), \textit{w/o} denotes without using SMPL body priors, \textit{Can.} or \textit{Obs.} means using  body SDF priors from canonical or observation spaces, \textit{Obs.}+raw means predicting raw SDFs with priors from observation spaces.
	}
	\label{ablations}
	\vspace{-5mm}
\end{table*}

\begin{figure*}[t]
    \centering
    \small
    \includegraphics[width=0.85\linewidth]{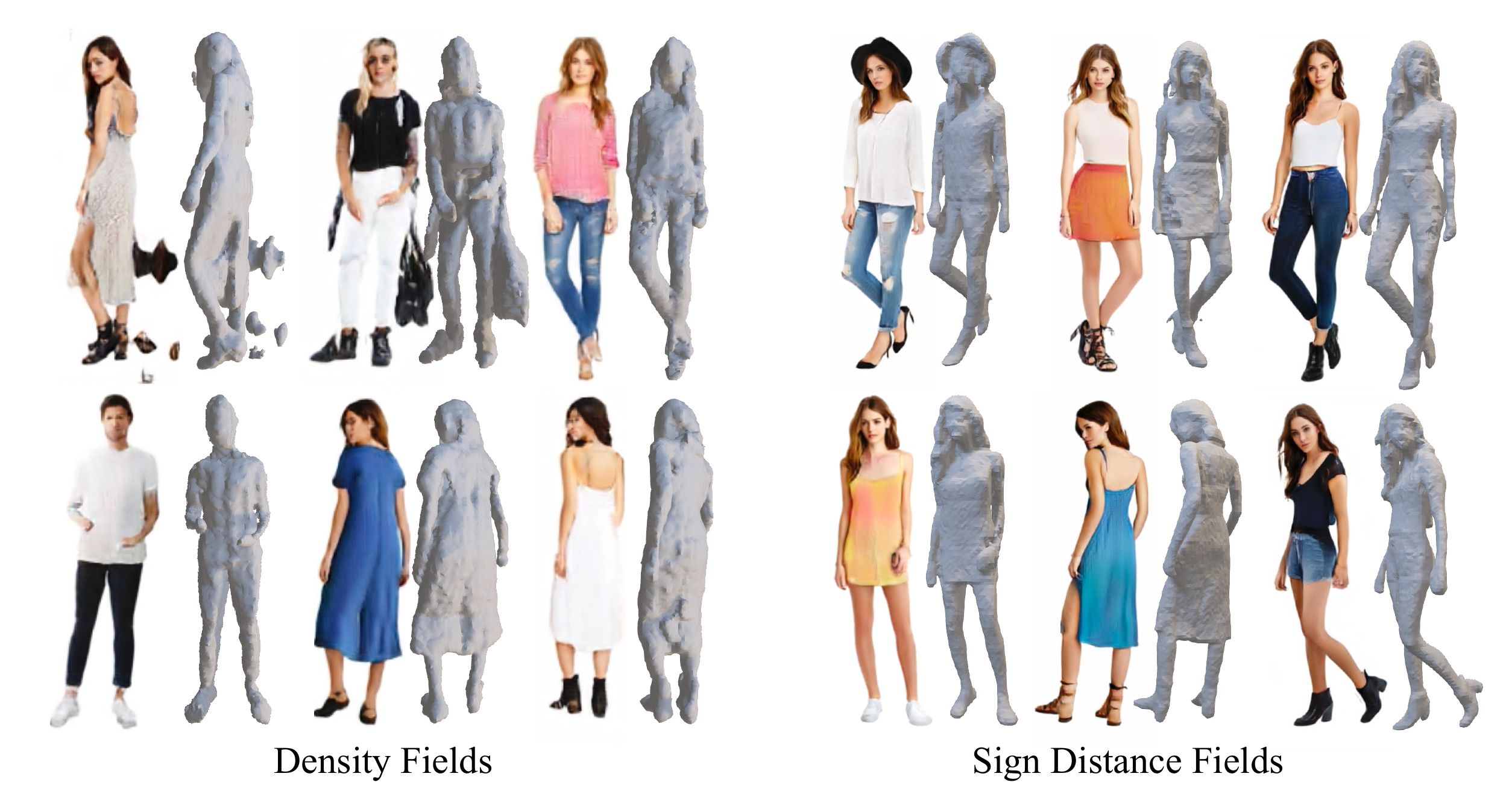}
    \caption{\textbf{Ablation on geometry proxy.} We compare the results of \nameofmethod{} trained with density and signed distance fields.} 
    \label{fig:density}
    \vspace{-4mm}
\end{figure*}

\begin{figure*}[h]
    \centering
    \small
    \includegraphics[width=0.85\linewidth]{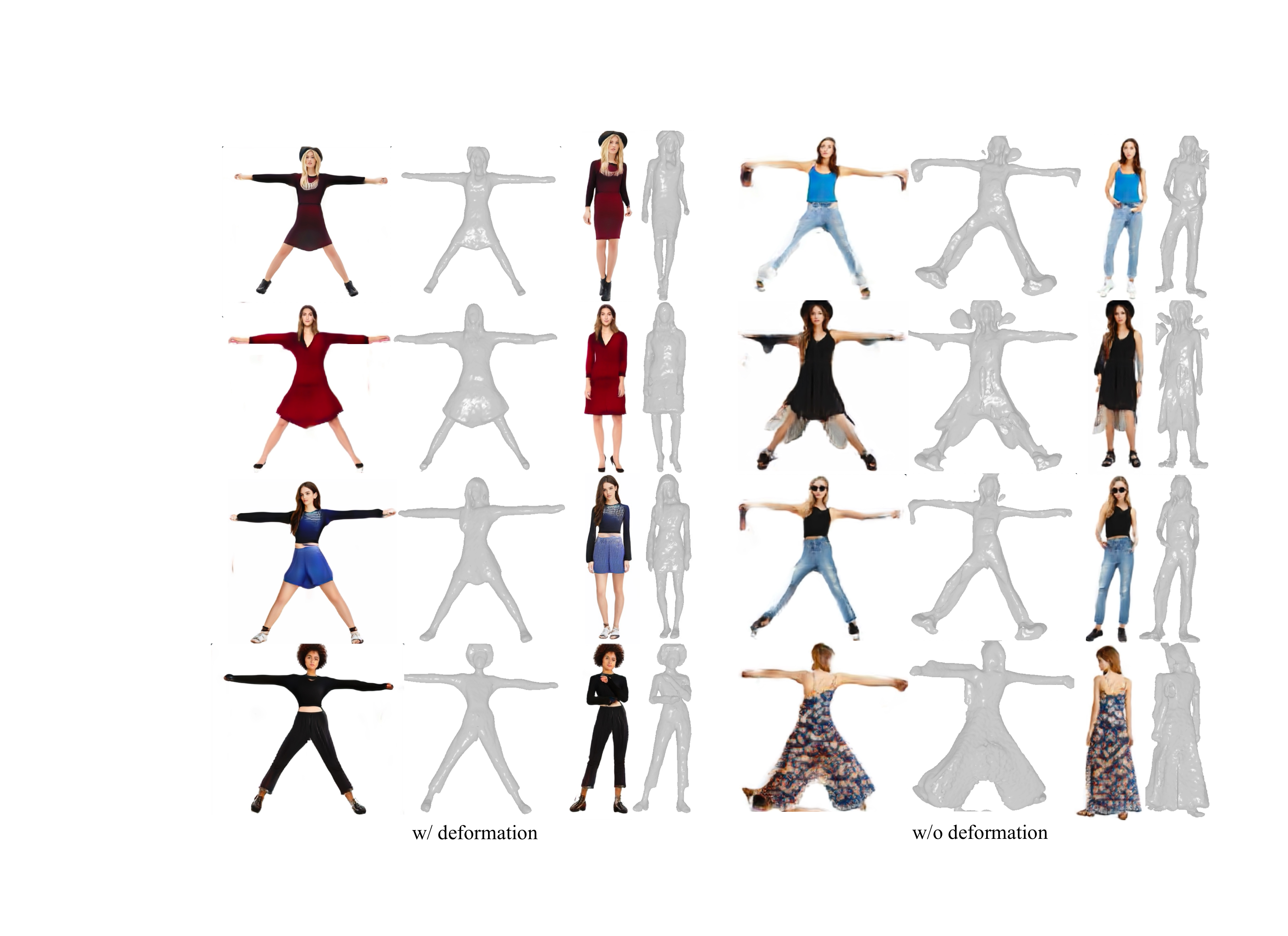}
    \caption{\textbf{Ablation on deformation network.} The visualization of the generated canonical and  posed avatars of our model trained with (left) and without (right) deformation fields. Clear artifacts can be observed if we remove the deformation module during model training.} 
    \label{fig:deformation}
\end{figure*}

\subsection{Ablation studies}
\label{sec:ablations}

We conduct ablation studies on DeepFashion ($256\times 256$) as it contains more  diverse appearances and poses.

\vspace{1mm}
\noindent  \textbf{Geometry proxy.} 
Our \nameofmethod{} uses SDF as geometry proxy to facilitate the geometry learning. To study its effectiveness, we  also evaluate our model with density field as the proxy. As shown in Tab.~\ref{ablation:geoproxy}, using the density field degrades     quality of the generated avatars   significantly\textemdash 16.8\% and 22.0\% increases in FID and FaceFID, and 11.6\% increase in Depth error. This indicates SDF is important for the model to more precisely represent clothed human geometry, especially for detailed regions like face. 

We also show quantitative results of the model trained with density (left) or SDF (right) as geometry proxy in Fig.~\ref{fig:density}.
We observe the model trained with density fields can generate  human avatars with reasonable geometry. However, it tends to generate noisy  body surface. Besides, redundant geometry (\eg, first two examples) is observed in the background regions due to the lack of geometric constraints. Compared with it, the model trained with SDF yields smoother human body surface with better quality thanks to the geometric regularization introduced by SDF.

\vspace{1mm}
\noindent  \textbf{SDF prediction schemes.}  
\nameofmethod{} 
predicts residual SDFs of clothed humans on top of the  SMPL body mesh. As shown in Tab.~\ref{ablation:bodysdf}, if removing SMPL body guidance and directly predicting residual SDFs (1st row), the performance drops significantly, \ie, 8.7\%, 17.8\%, 16.4\% increase in FID, FaceFID and Depth error. This indicates the coarse SMPL body prior is important for guiding \nameofmethod{} to better generate clothed human geometry. 
We also compare the performance between SMPL body SDFs queried from observation (\textit{Obs.}) and canonical (\textit{Can.}) spaces. The model guided by the body SDFs queried from Obs., which are more accurate,  yielding better performance.

In addition, we study the effect of two SDF prediction schemes\textemdash predicting raw SDFs directly (\textit{Obs.+raw}) and predicting SDF residuals on top of the coarse SMPL body from observation space. We see the residual prediction scheme delivers better results as it alleviates the geometry learning difficulties.
Moreover, we study the effect of SMPL geometric prior loss in Eqn.~(\ref{eqn:prior}). Removing it will lead to   performance drop w.r.t.\  all the metrics   (FID: 7.20 \emph{vs.} 6.54, FaceFID: 10.29 \emph{vs.} 8.74, Depth: 0.645 \emph{vs.} 0.576), verifying its effectiveness  for regularizing geometry learning.

\vspace{1mm}
\noindent  \textbf{Deformation schemes.} 
Our model uses a pose-guided deformation to un-warp spatial points from the observation  space to the canonical space. We also evaluate  other two deformation schemes in Tab.~\ref{ablation:deformation}: 1) residual deformation~\cite{park2021nerfies} only (\textit{RD}), 2) inverse LBS deformation~\cite{huang2020arch} only (\textit{IS}). 
When using RD only, the model training does not converge, indicating that learning deformation implicitly cannot handle large articulation of humans and   lead to implausible results.
When using IS only, the model achieves a reasonable result (FID: 7.89, FaceFID: 12.9, Depth: 0.752), verifying the importance of the explicitly pose-guided deformation. 
Further combining IS and RD  (our model) boosts the performance sharply\textemdash 20.6\%, 32.2\% and 23.4\% decrease in FID, FaceFID and Depth respectively. 
These results show  the residual deformation, collaborating with the posed-guided inverse LSB transformation, indeed  better represents non-rigid clothed human body deformation, yielding better appearance and geometry modeling.

To better study the effects of the proposed deformation network, we visualize generation results (\ie, avatars in the canonical and  target observation spaces) of \nameofmethod{} trained with (left) and without (right) it in Fig.~\ref{fig:deformation}. From the figure, we  observe the deformation network helps to learn better canonical to observation mapping and clearly improves the quality of the generated canonical avatars, and thus lead to high-quality posed avatars generation. Notably, our deformation network can help to build the correct correspondence between the canonical to observation spaces even for loose clothes like dresses.

\vspace{1mm}
 \noindent  \textbf{Face discriminator.} Tab.~\ref{ablation:facedisc} shows the effect of using the face discriminator. We can observe a significant drop in the quality of faces if disable the face discriminator, \ie, 12.31 \emph{vs.} 8.74 in FaceFID. Although the overall FID is slightly improved, the geometry quality becomes worse. This indicates the additional face discriminator is helpful for   learning     geometry details and improving    quality of the target region. Please see qualitative comparisons in Appendix.

\vspace{1mm}
 \noindent  \textbf{Number of ray steps.}  Tab.~\ref{ablation:ray_step} shows the effect of the number of points sampled per camera ray. With only 12 sampled points for each ray, \nameofmethod{} already achieves acceptable results, \ie, 7.76, 12.44 and 0.656 in FID, FaceFID and Depth. With more sampling points, the performance monotonically increases, demonstrating  the  capacity of \nameofmethod{} in 3D-aware human synthesis.

 \vspace{1mm}
\noindent \textbf{Number of KNN in inverse skinning deformation.}
For any spatial points, we use Nearest Neighbor to find the corresponding skinning weights for the inverse LBS transformation (Eqn.~(\ref{eqn:invskin})). Here we study how the number of KNN neighbors  affects the model performance in Tab.~\ref{ablation:knn}. We observe that using more KNN neighbors gives slightly better FID performance, but worse geometry and much longer training time as the model need to retrieve more neighbors.

\label{sec:applications}

\begin{figure}[t]
    \centering
    \small
    \includegraphics[width=\linewidth]{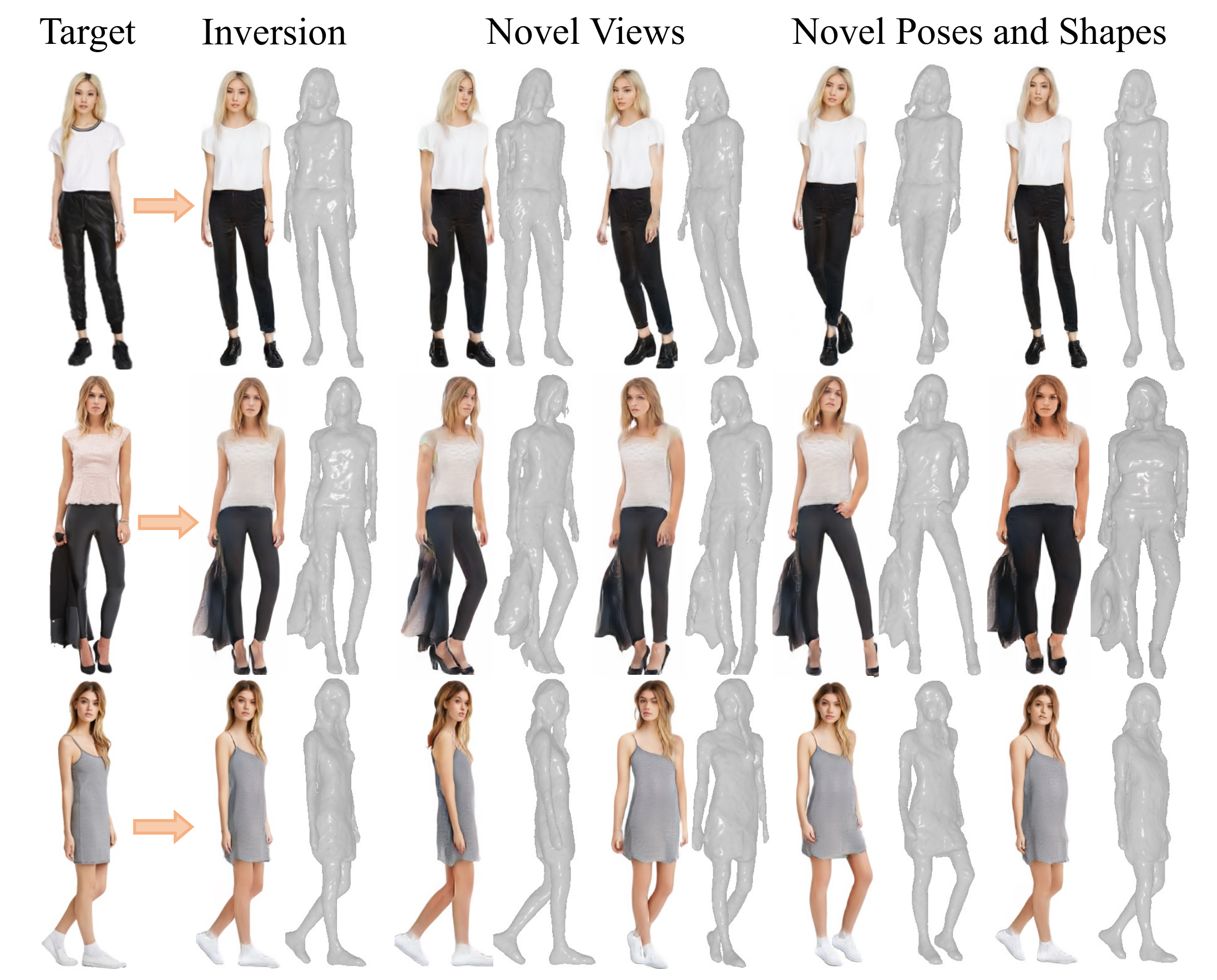}
 \caption{ \textbf{Portrait image reconstruction and animation} results of \nameofmethod{}. Given target portrait image, we reconstruct 3D-aware appearance and geometry of the human, who is rendered under novel camera views, and re-posed/shaped by taking novel SMPL parameters as control signals and animated accordingly. The model for inversion is trained with DeepFashion and the target images are from SHHQ. Better viewed in $2\times$ zoom.
 } 
    \label{fig:inversion}
\end{figure}

\subsection{Applications}

\noindent  {\bf Portrait image reconstruction and animation.}
Fig.~\ref{fig:inversion} shows the application of \nameofmethod{} for single-view 3D reconstruction. {Following~\cite{song2021agilegan}, we fit the target images by optimizing the latent code with MSE loss and perceptual loss to recover both the appearance and  geometry. With the recovered 3D representation, we manipulate the images under novel camera poses. Thanks to its disentangled design, \nameofmethod{} also can re-animate the generated avatars with a preserved identity and high-fidelity textures under different poses and shapes specified by the SMPL control  signals.}

\begin{figure}[t]
    \centering
    \small
    \includegraphics[width=\linewidth]{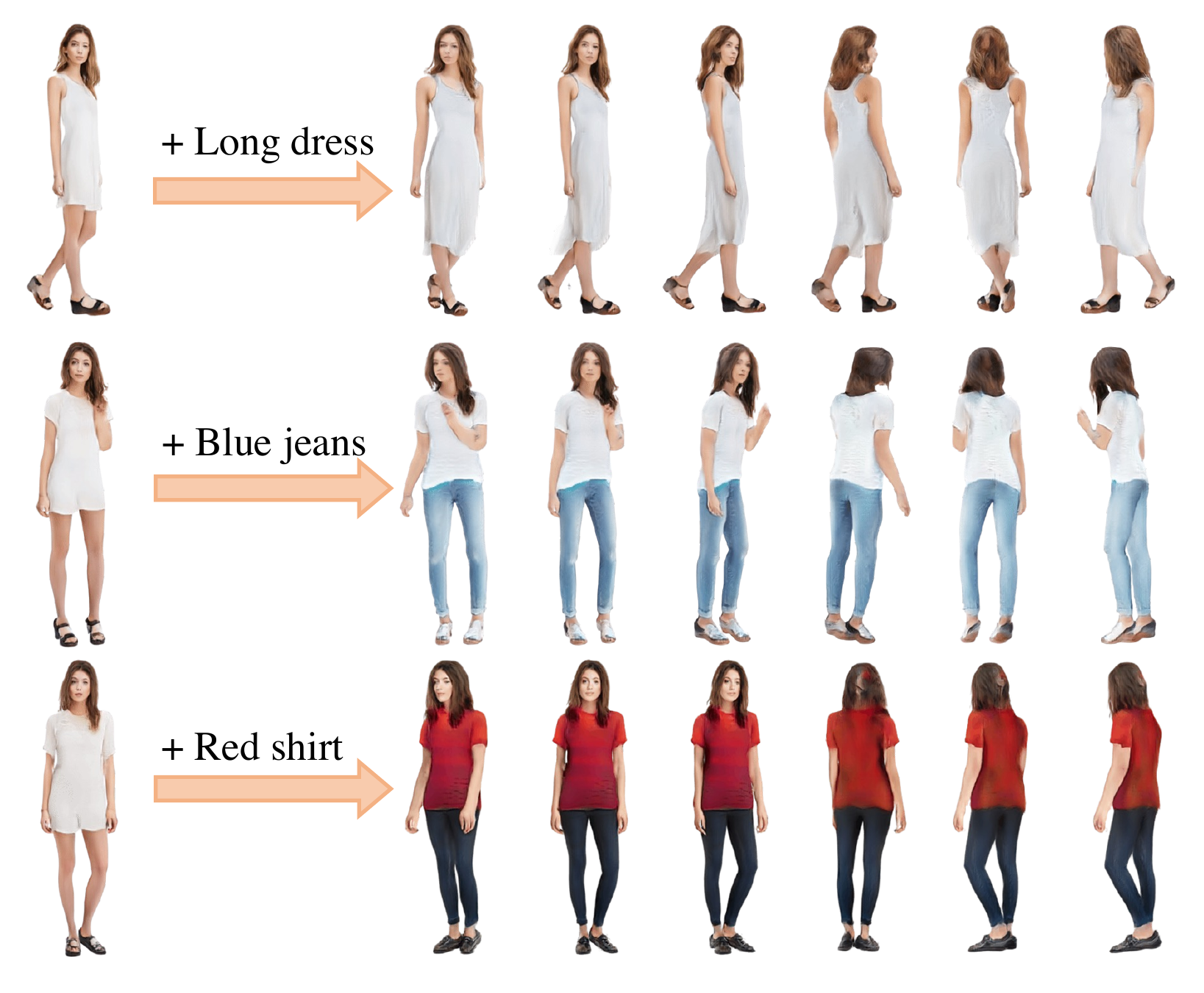}
 \caption{\textbf{Text-guided  synthesis} results of \nameofmethod{} with multi-view rendering. Different poses and text prompts are used for CLIP-guided optimization. Better viewed in $2\times$ zoom.} 
    \label{fig:clip_inversion}
\end{figure}

\noindent  {\bf Text-guided synthesis.}
Recent works~\cite{Patashnik_2021_ICCV,kwon2022clipstyler} have explored  using a text-image embedding~\cite{radford2021learning} to guide GANs for controllable image synthesis. We also visualize text-guided clothed human synthesis in Fig.~\ref{fig:clip_inversion}. Specifically, we follow StyleCLIP~\cite{Patashnik_2021_ICCV} to optimize the latent code of the synthesized images with a sequence of text prompts that specify different cloth styles. 
Then we synthesize the re-dressed human avatars under different camera poses.

\section{Conclusion}
\label{sec:conclusion}

This work introduced the first 3D-aware clothed human avatar generative model, \nameofmethod{}. By factorizing the generative process into the canonical human generation and deformation stages, it  can leverage the  geometry priors and  effective tri-plane representation to  address  the challenges in animatable human avatar generation.  We demonstrated it can generate high-fidelity humans with disentangled geometry controllability, and support several downstream applications. 
We believe our method will make the creation of human avatars more accessible to ordinary users, assist designers  and reduce the manual cost.  

\noindent  {\bf Limitations.} 
We presented a high-fidelity animatable 3D human GAN, but there is still space for further improvement in our method.
1) It relies on  the SMPL estimations. Inaccurate    estimations   would lead to generation quality  degradation.
2) The fine-grained motions of the human avatars  generated by our method,  like face expression changes,  cannot be controlled yet. Using more expressive 3D human model, \eg, SMPL-X as guidance would be a promising solution.

{\small
\bibliographystyle{ieee_fullname}
\bibliography{egbib}
}


\clearpage
\appendix

\renewcommand{\thetable}{S\arabic{table}}
\renewcommand{\thefigure}{S\arabic{figure}}

\maketitle

We provide more implementation details in Section~\ref{sec:detail}. We present additional baseline comparisons, ablation studies and more visual results in Section~\ref{sec:analysis}. For  the results in video format, please refer to the \href{http://jeff95.me/projects/avatargen.html}{project page}.

\section{Implementation Details}
\label{sec:detail}

\noindent We detail the implementation of each module of  \nameofmethod{} and the hyper-parameter choice in this section.

\vspace{6pt} \noindent \textbf{Backbone.} We adopt the StyleGAN2~\cite{Karras2020stylegan2} backbone as tri-plane generator and two blocks
of StyleGAN2-modulated convolutional layers as super-resolution module, following EG3D~\cite{chan2021efficient}. 
The output tri-plane resolution is set as $256\times 256$ with 96 channels which are then split to three planes for feature sampling. 
We use the MLP-based mapping network in StyleGAN2 to encode the latent code and camera condition.
We adopt a similar dual-discriminator architecture used in EG3D for regularizing the generator training. Different from EG3D, we further feed an additional upsampled image of the head region to the discriminator to improve the generation quality of human faces. For $512^2$-resolution image, an $80^2$-squared image covering the whole head region is cropped for face discrimination.

\begin{figure}[h]
    \centering
    \includegraphics[width=\linewidth]{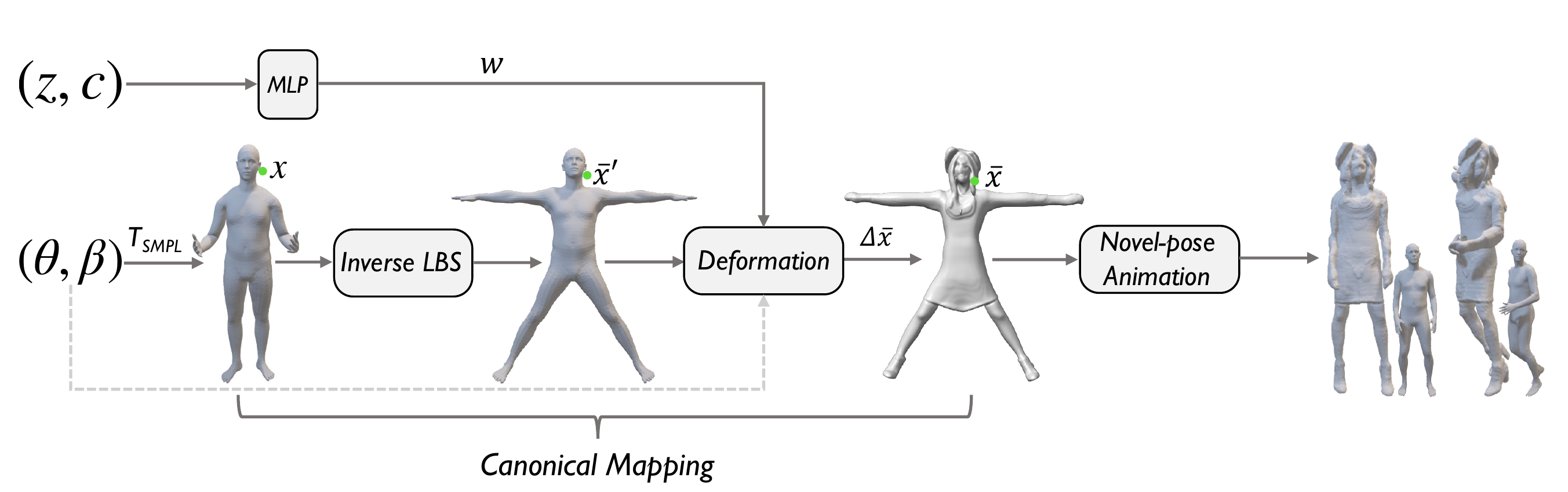}
    \caption{Detailed architecture of our canonical mapping.} 
    \label{fig:canonical}
\end{figure}

\vspace{6pt}  \noindent \textbf{Canonical mapping.} Fig~\ref{fig:canonical} shows   details   of the canonical mapping module. For a  point  $\bb{x}_i$, we first find its nearest point $\bb{v^{*}}$ on the SMPL body surface: $\bb{v^{*}}=\argmin_{\bb{v}\in \mathcal{V}}||\bb{x}_i-\bb{v}||_2$. 
Then, we use its corresponding skinning weights $\bb{s^*}$ to deform $\bb{x}_i$ to $\bar{\bb{x}}_i'$ in the canonical space as:
\begin{equation}
    \bar{\bb{x}}_i' = T_o (\bb{x}_i| \bb{p}) = T_{\text{IS}} ({\bb{x}_i}, \mathbf{s^*}, \bb{p}).
    \label{eqn:invskin_sup}
\end{equation}
For the fine-grained geometric deformation, we further train a deformation network to learn a residual deformation:
\begin{equation}
\begin{aligned}
   \Delta \bar{\bb{x}}_i = \text{MLPs}(\text{Concat}[\text{Embed}(\mathbf{x}_i),\mathbf{w},\mathbf{p}]),
\end{aligned}
\end{equation}
where $\mathbf{w}$ is the latent style code mapped from the input latent code $\mathbf{z}$ via MLP, which contains appearance and geometric details of the current generation. We concatenate it with the position-embedded $\mathbf{x}_i$ and SMPL $\mathbf{p}$ and feed them to MLPs to yield the residual deformation $\Delta \bar{\mathbf{x}}_i$. 
Thus, the final canonical mapping $T$ is formulated as
\begin{equation}
\begin{aligned}
    \bar{\bb{x}}_i = T(\mathbf{x}_i)  =\bar{\bb{x}}_i'+\Delta \bar{\mathbf{x}}_i.
\end{aligned}
\end{equation}

During inference, given the latent code $\bb{z}$ and camera $\bb{c}$, \nameofmethod{} leverages the pre-trained tri-plane to synthesize the corresponding appearance and geometry of the canonical avatar, and then animate it by deforming the canonical one according to the desired SMPL control signals.

\begin{figure}[!h]
    \centering
    \includegraphics[width=\linewidth]{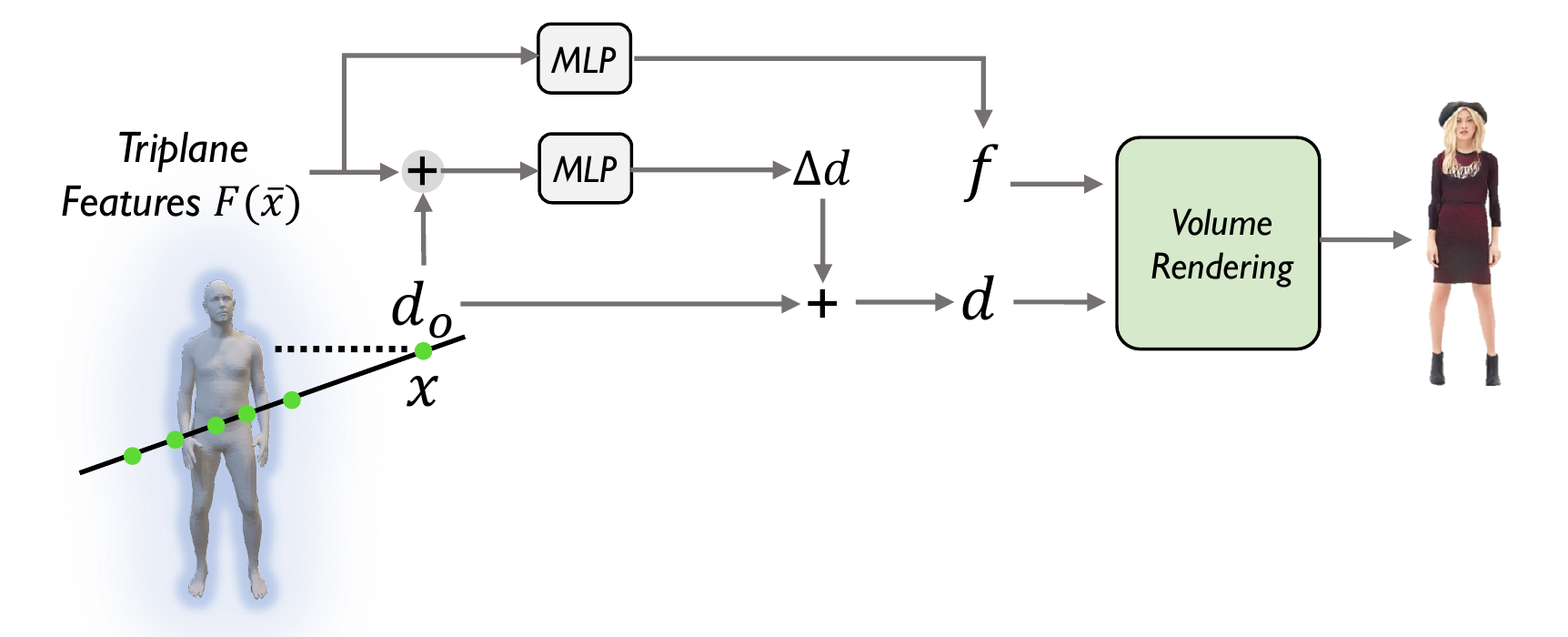}
    \caption{Detailed architecture of SDF-based volume renderer.} 
    \label{fig:sdf_renderer}
\end{figure}

\vspace{6pt}  \noindent \textbf{SDF-based volume renderer.}
The detailed architecture of our SDF-based volume renderer is shown in Fig.~\ref{fig:sdf_renderer}. 
For any sampled points ${\bb{x}}$ on the rays, we  first un-warp it  to $\bar{\bb{x}}$ by canonical mapping (Fig.~\ref{fig:canonical}). We query feature $F(\bar{\bb{x}})$ for $\bar{\bb{x}}$ from the canonical tri-plane, and  feed it into two MLPs to predict the color feature $\bb{f}=\text{MLP}_f(F(\bar{\bb{x}}))$ and the signed distance $d=d_o+\Delta d=d_o + \text{MLP}_{d}(F(\bar{\bb{x}}), d_o)$, where $d_o$ is body SDF queried from SMPL for ${\bb{x}}$. 
We then convert the signed distance value $d_i$ of each point $\mathbf{x}_i$ along a ray $R$ to density value $\sigma_i$ as
$
\sigma_i = \frac{1}{\alpha}\cdot \text{Sigmoid}(\frac{-d_i}{\alpha}),
$
where $\alpha >0$ is a learnable parameter that controls the tightness of the density around the surface boundary.
By integration along the ray $R$ we can get the corresponding pixel feature as 
\begin{equation}
    I(R) = \sum_{i=1}^N \left(\prod_{j=1}^{i-1} e^{-\sigma_j \cdot \delta_j}\right) \cdot \left(1-e^{-\sigma_i \cdot \delta_i}\right) \cdot \bb{f}_i,
\end{equation}
where $\delta_i = || \bb{x}_i-\bb{x}_{i-1} ||$ and $N$ is number of points sampled per ray. By aggregating all rays, we can get the high-dimensional feature image for the following super-resolution module.

\vspace{6pt}  \noindent \textbf{Hyper-parameter.} The learning rate is set as $2.5\times 10^{-3}$ for generator and $2\times 10^{-3}$ for discriminator. For loss function, we set $\lambda_{\text{Reg}}=10$, $\lambda_{\text{Eik}}=10^{-3}$, $\lambda_{\text{Minsurf}}=0.05$, $\lambda_{\text{Deform}}=10$, $\lambda_{\text{Prior}}=1$. The model is trained on 8 NVIDIA V100 GPUs with batch size of 16. Similar to EG3D, we adopt a progressive growing strategy, \ie, we first train the model with $64^2$-resolution volume rendering and then progressively grow it to $128^2$-resolution. The total training process takes about 4 days to converge.

\section{More results}
\label{sec:analysis}

\subsection{Face discriminator}
\begin{figure*}[!h]
    \centering
    \includegraphics[width=0.95\linewidth]{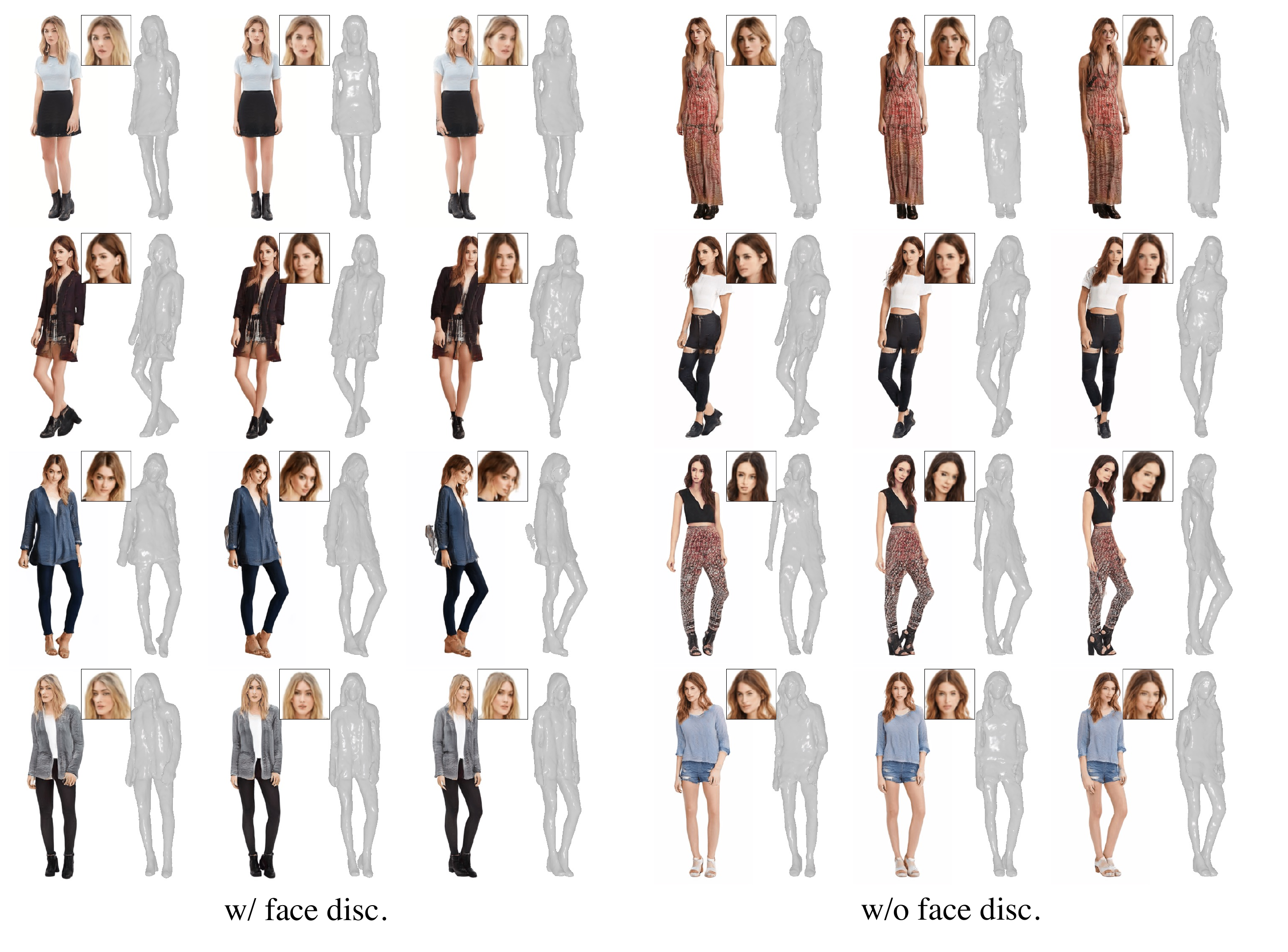}
    \caption{Ablation on face discrimination, the result generated by our model with face discrimination is shown on the left side and the result without face discrimination is shown on the right side. Multi-view rendering of both geometry and appearance are listed. Face regions are cropped and zoomed in for better visualization.} 
    \label{fig:face}
\end{figure*}

We show the comparison of our model trained with and without face discrimination in Fig.~\ref{fig:face}. We can observe that our proposed face discrimination clearly improves the quality of the face region. If we disable the face discrimination, the model tends to produce distorted faces especially when rendered from different camera poses.

\subsection{Disentanglement between geometry and appearance}
\begin{figure*}[h]
    \centering
    \includegraphics[width=0.7\linewidth]{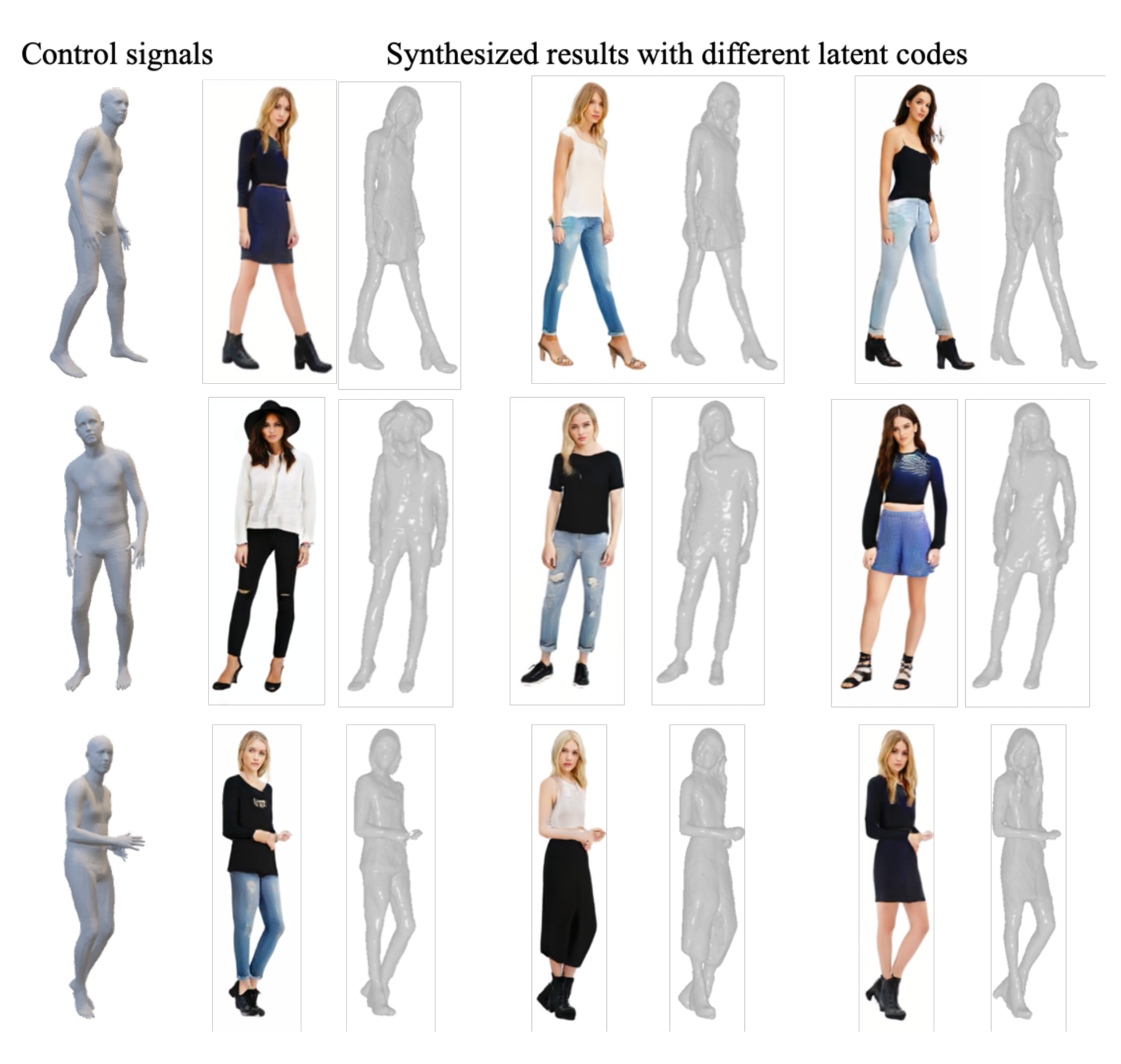}
    \caption{Avatar generated under the same SMPL condition but different latent codes.} 
    \label{fig:disentanglement}
\end{figure*}

In Fig.~\ref{fig:disentanglement}, we visualize more synthesized human avatars with the same pose and shape conditions but different latent codes.
We observe \nameofmethod{} can synthesize diverse avatars with  the desired shape and pose as specified by the input SMPL signals. This clearly demonstrates the proposed decomposed pipeline can well disentangle geometry generation from appearance, making animatable avatar generation possible.

\subsection{Comparison with 2D GAN}
\begin{figure*}[h]
    \centering
    \includegraphics[width=0.7\linewidth]{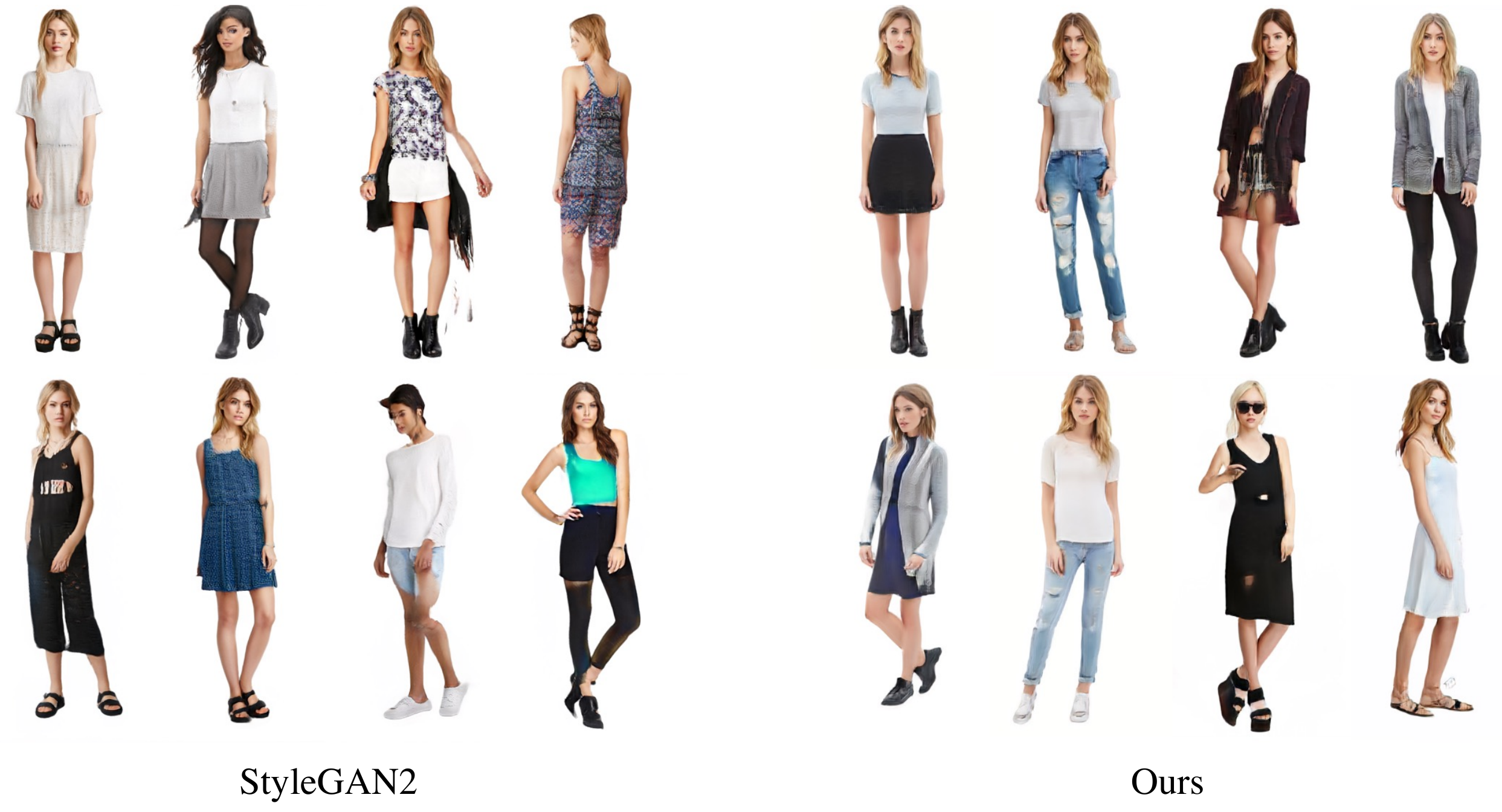}
    \caption{Comparison of 2D StyleGAN2 with our \nameofmethod{}.} 
    \label{fig:2d_gan}
\end{figure*}

We show the comparison of 2D StyleGAN2 with our generation results in Fig~\ref{fig:2d_gan}. Both models are trained on DeepFashion with $512^2$ resolution. Although the FID of  StyleGAN2 is much lower than ours (\ie, 3.07 \vs 7.68), it is difficult for StyleGAN2 to capture the structure of  human body and thus  suffers from the artifacts of unnatural poses and distorted body parts. Our method, however, with the geometry priors and  learned 3D-aware representation, can effectively capture the structure of  human body and generate high-quality images  given the desired condition.

\subsection{More visualization results}
\begin{figure*}[!t]
    \centering
    \includegraphics[width=\linewidth]{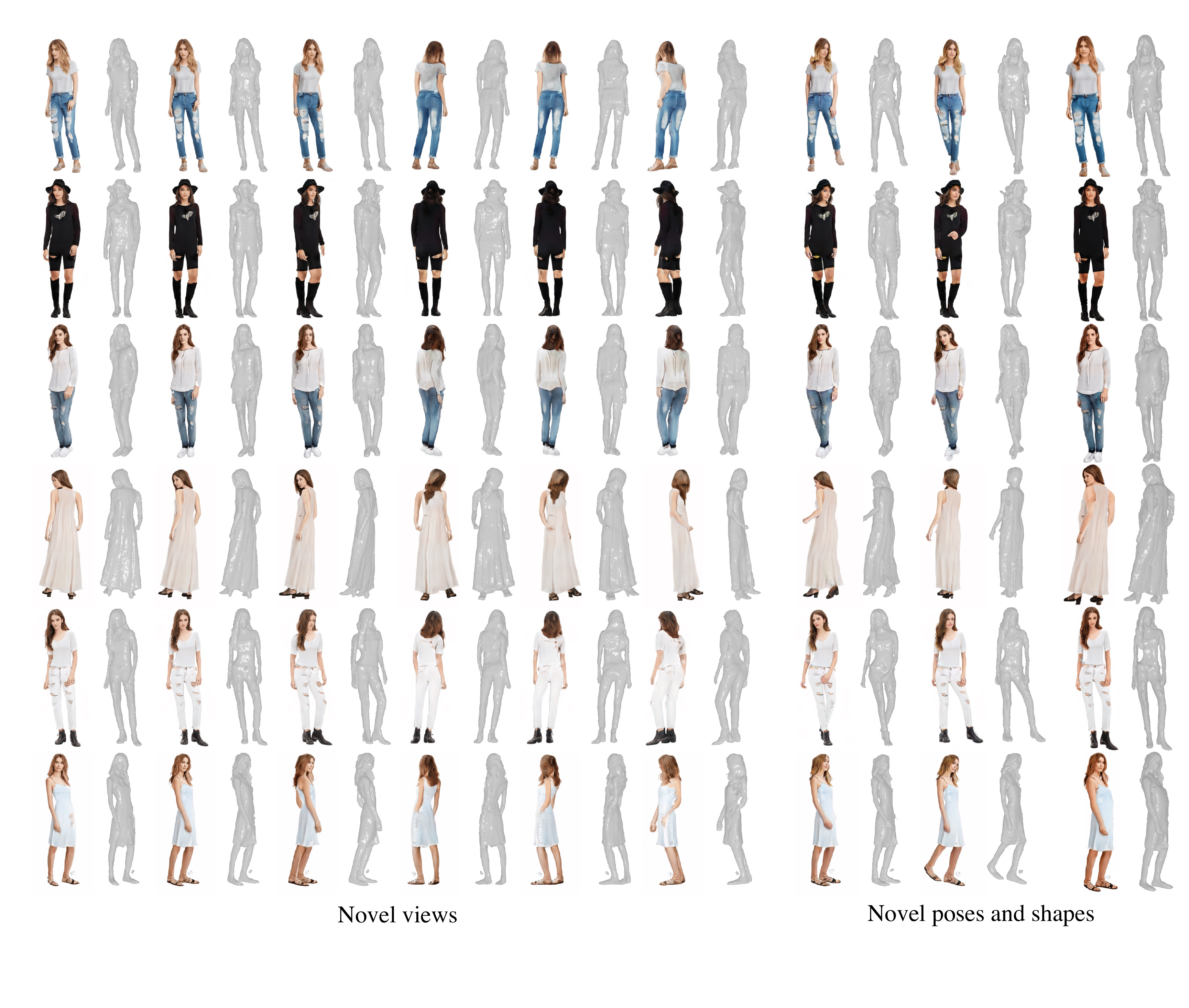}
    \caption{More multi-view rendering and novel pose/shape synthesis results. Best viewed in $2\times$ zoom.}
    \label{fig:multiview}
    \vspace{-10mm}
\end{figure*}

\begin{figure*}[h]
    \centering
    \includegraphics[width=\linewidth]{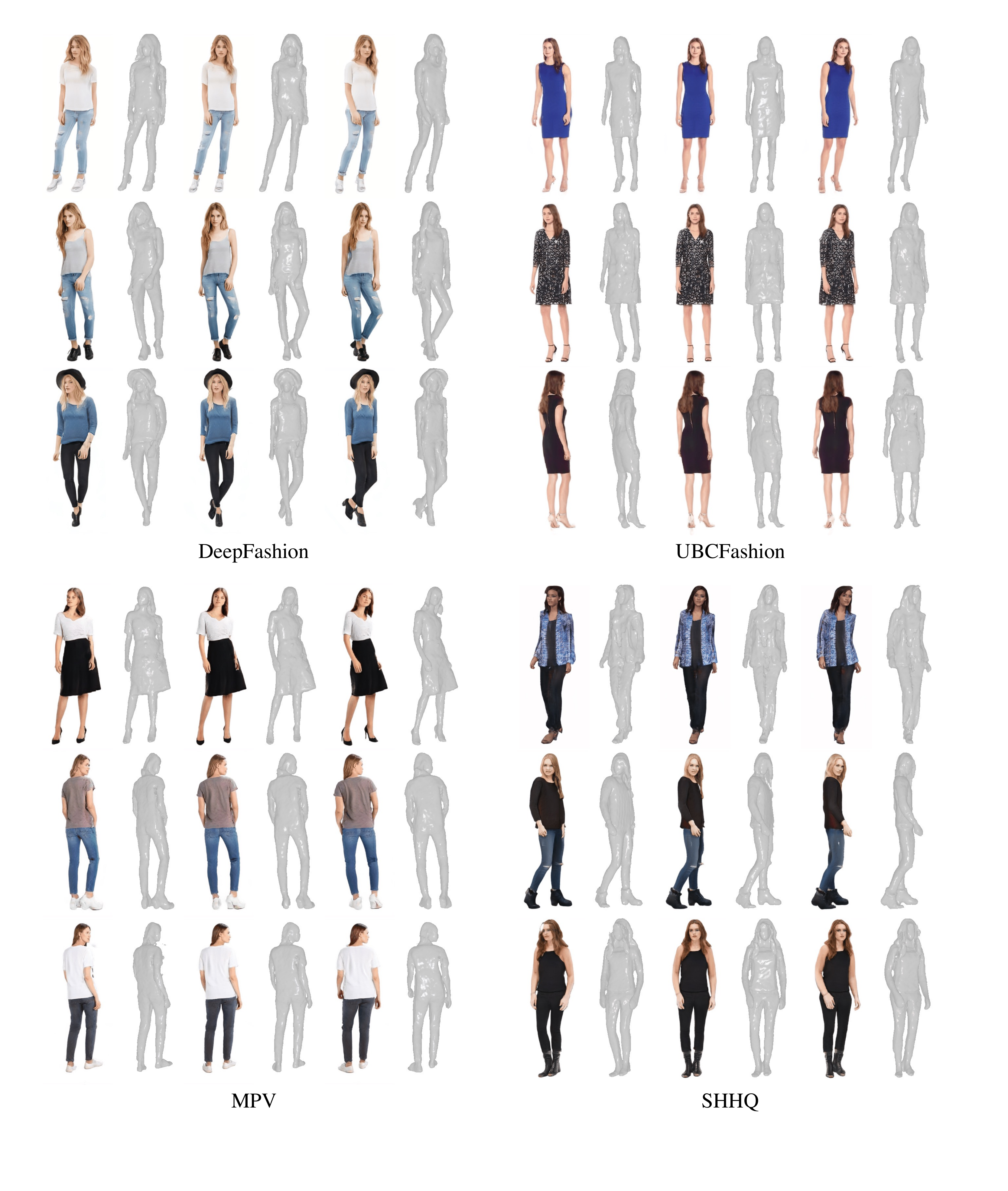}
    \caption{More multi-view rendering results of \nameofmethod{} on four datasets including DeepFashion~\cite{liuLQWTcvpr16DeepFashion}, UBCFashion~\cite{zablotskaia2019dwnet}, MPV~\cite{dong2019towards} and SHHQ~\cite{fu2022styleganhuman}.}
    \label{fig:dataset}
\end{figure*}
\begin{figure*}[h]
    \centering
    \includegraphics[width=0.9\linewidth]{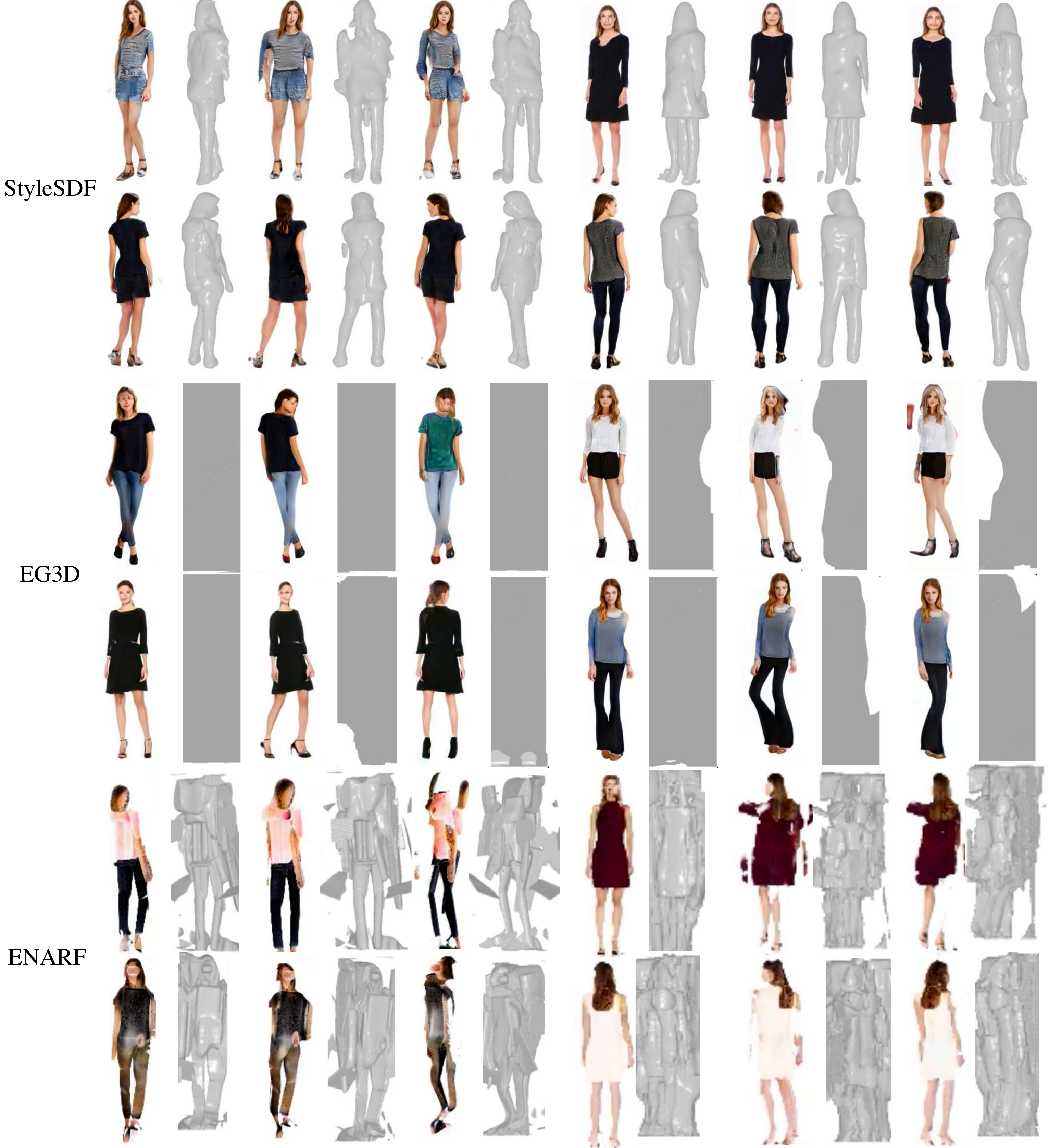}
    \caption{More multi-view rendering results of baseline methods: StyleSDF~\cite{or2021stylesdf}, EG3D~\cite{chan2021efficient} and ENARF~\cite{noguchi2022unsupervised}.}
    \label{fig:baseline_comp}
\end{figure*}

We show more visualization results of our method in Fig.~\ref{fig:multiview} and Fig.~\ref{fig:dataset}, as well as multi-view rendering results of baseline methods on Fig.~\ref{fig:baseline_comp}.
Please visit our \href{http://jeff95.me/projects/avatargen.html}{project page} for results in video format.

\end{document}